\documentclass{article}

\PassOptionsToPackage{numbers, ,sort&compress}{natbib}

\usepackage[main, final]{neurips_2026}

\usepackage[utf8]{inputenc} 
\usepackage[T1]{fontenc}    
\usepackage{hyperref}       
\usepackage{url}            
\usepackage{booktabs}       
\usepackage{amsfonts}       
\usepackage{nicefrac}       
\usepackage{microtype}      
\usepackage{xcolor}         

\usepackage{float}         
\usepackage{graphicx}
\usepackage{enumitem}
\usepackage{wrapfig}
\usepackage{amsmath}
\usepackage{multirow}
\usepackage{tabularx}
\usepackage{adjustbox}
\usepackage{needspace}

\definecolor{editinglavender}{RGB}{226,218,245}
\definecolor{steeringpeach}{RGB}{253,221,199}

\newcommand{\editingmethod}[1]{%
  \begingroup
  \setlength{\fboxsep}{1pt}%
  \colorbox{editinglavender}{\vphantom{Ay}#1}%
  \endgroup
}

\newcommand{\steeringmethod}[1]{%
  \begingroup
  \setlength{\fboxsep}{1pt}%
  \colorbox{steeringpeach}{\vphantom{Ay}#1}%
  \endgroup
}

\usepackage{caption}
\usepackage{wrapfig}

\usepackage{subcaption}
\usepackage{etoc}           
\usepackage{placeins}       

\definecolor{linklavender}{RGB}{150,92,190}
\hypersetup{
    colorlinks=true,
    citecolor=linklavender,
    linkcolor=linklavender,
    urlcolor=linklavender
}

\definecolor{appred}{RGB}{200,40,50}
\newcommand{\appref}[1]{{\hypersetup{linkcolor=appred}\ref{#1}}}
\title{Few-Shot Visual Concept Extraction for Steering Diffusion Transformers}

\author{%
  Nabyl Quignon \\
  Inria Centre at Université Côte d'Azur\\
  \texttt{nabyl.quignon@inria.fr} \\
  \And
  Antitza Dantcheva \\
  Inria Centre at Université Côte d'Azur \\
  \texttt{antitza.dantcheva@inria.fr} \\
}

\begin{document}

\maketitle

\begin{abstract}
Image steering aims to provide fine-grained control over concepts in generated images, enabling targeted and continuous manipulation without broadly altering the generation process. Recent steering methods for diffusion models primarily focus on the text-conditioning space or rely on costly auxiliary training. Deviating from that, we here propose a novel steering approach that extracts visual concept directions directly from the image-side activations of Diffusion Transformers. Specifically, we showcase few-shot concept extraction in the activation space using a simple difference-of-means (DoM) estimator over intermediate residuals. We observe (a) scene-level concepts such as lighting, weather, and style are distributed across image tokens, as well as (b) local concepts such as object colors, textures, materials, and facial attributes are concentrated in specific regions. Towards controlling both concepts, we introduce a new unified training-free framework that uses global difference-of-means for (a) distributed concepts and masked difference-of-means for (b) local concepts, preventing local signals from being diluted by global averaging. Notably, using only a handful of positive and negative prompt pairs, our method identifies and extracts reusable concept directions that can be injected at inference for continuous steering. Quantitatively, our method achieves substantially stronger edits than text-encoder steering and training-free activation-editing baselines, and approaches trained reference-image editing models. Our results suggest that as concept directions can be applied at patch level, our method enables targeted and compositional steering, manipulating multiple concepts in the same image. Project page: \href{https://newbyl.github.io/few_shot_project_page/}{\nolinkurl{newbyl.github.io/few_shot_project_page}}.
\end{abstract}

\section{Introduction}

Diffusion Transformers (DiTs) have emerged as a powerful foundation for image generation~\citep{esser2024scalingrectifiedflowtransformers, flux-2-2025, flux2024, wu2025qwenimagetechnicalreport}. However, achieving \textit{fine-grained control over} specific visual \textit{concepts} in their outputs remains challenging. Many targeted image modifications are not purely semantic but rather visual and spatial in nature, including altering object color, modifying material properties, adjusting facial expressions, or shifting global scene appearance while preserving structural content. Given that such attributes are inherently encoded in the image-side representations of the generator, activation-space interventions bring to the fore an alternative approach to prompt-only control~\citep{parihar2025kontinuouskontext, zarei2025slideredit}.

Existing approaches only partially address the challenge of fine-grained visual control in Diffusion Transformers. Some methods place emphasis on \textit{steering in the text-conditioning} space~\citep{kamenetsky2025saedittokenlevelcontrolcontinuous}, providing a lightweight steering interface,  requiring large sets of positive and negative prompt pairs, while never directly accessing the image tokens where visual attributes are instantiated. Others rely on \textit{auxiliary training} or \textit{edit-specific modules}~\citep{zarei2025slideredit, parihar2025kontinuouskontext}, which can provide strong control that comes at the cost of additional data, optimization, or model-specific training. These limitations motivate a fundamental question: \textit{can visual concept directions be extracted directly from DiT activations using only a few examples and without auxiliary training?}

In this paper, we showcase that \textit{direct extraction of visual concept directions} is indeed feasible. Towards that we adopt the \textit{difference-of-means estimator} \citep{rimsky-etal-2024-steering, li2024inference, arditi2024refusal}: given positive and negative examples of a concept, we estimate a steering direction by averaging the difference between their intermediate activations. This simple estimator proves effective for \textit{global concepts} such as lighting, weather, style, or scene appearance, where averaging over all image tokens constitutes meaningful directions due to global concept representation.

\begin{figure}[H]
    \centering
    \includegraphics[width=\textwidth]{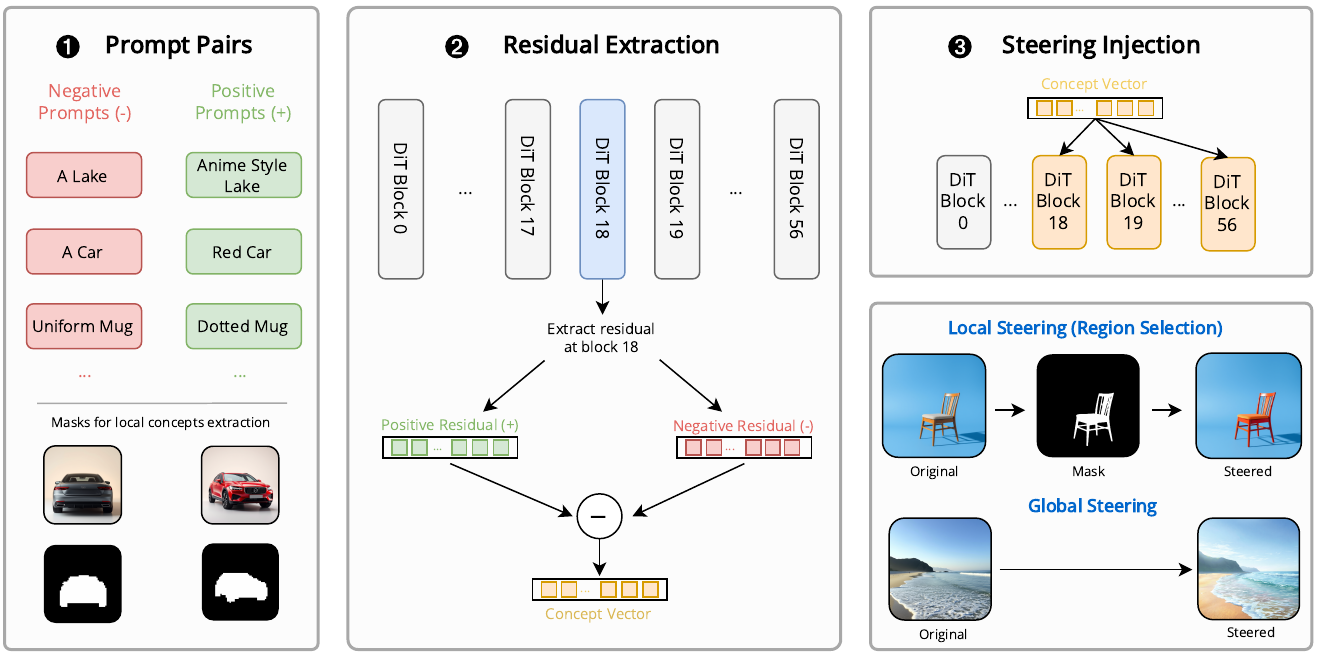}
    \vspace{-0.4cm}
    \caption{\textbf{Overview of our few-shot concept extraction and steering pipeline.} (1) We start from a small set of negative and positive prompt pairs, optionally with masks for localized concepts extraction. (2) We extract residual activations from a selected Diffusion Transformer block and estimate a concept direction using difference-of-means between positive and negative residuals. For global concepts, this is computed over all image tokens and for local concepts, it is restricted to the masked region patches. (3) At inference time, the resulting concept vector is injected into transformer blocks to steer generation, either globally or within a selected region.}
    \label{fig:teaser}
\end{figure}

However, global averaging is insufficient for \textit{local concepts} including object color, material texture, or facial attributes,  averaging over all image tokens as above dilutes the relevant signal with background, pose, layout, and features from unrelated objects. This failure is not attributed to activation-space extraction, but rather a spatial mismatch between the estimator's support and the concept's support. We address this mismatch with \textit{masked difference-of-means}. Given five positive and negative prompt-image pairs, we generate example images, localize the region where the target concept appears, and compute the activation difference exclusively over the corresponding image tokens. 

We achieve manipulation of both, global and local concepts in a \textit{unified framework for few-shot concept extraction in DiTs} that incorporates global difference-of-means for concepts distributed across the image, and masked difference-of-means for concepts localized to specific regions. To the best of our knowledge, this work is the first to propose spatial masking within the difference-of-means estimator for activation-space control in DiTs. Extracted directions are injected at inference for continuous steering, requiring no text encoder modification or auxiliary training. Across experiments global, object, and facial concepts, we find that a handful of positive and negative pairs suffice for effective control. As each concept direction is decoupled from its spatial region of application, our method enables targeted and compositional edits, such as applying different concepts to different objects within the same image.

Our contributions include the following.
\begin{itemize}
    \item We propose a novel image steering approach, namely a \textbf{training-free, few-shot framework} that extracts visual concept directions directly from intermediate DiT activations using five positive and negative prompt pairs with the difference-of-means estimator.
    \item We introduce a novel \textbf{masked difference-of-means}, a unified formulation that extracts global concepts from all image tokens and localized concepts from the tokens corresponding to the relevant region to address the limitations of naive difference-of-means.
    \item We show that \textbf{spatially localized concept extraction leads to more effective steering}: while global difference-of-means can capture broad scene-level concepts, masked difference-of-means is better suited to localized object and facial attributes, yielding stronger target edits while limiting unwanted drift from irrelevant spatial regions.
    \item We showcase that the extracted directions enable \textbf{continuous, targeted, and compositional steering} at inference, without modifying the text encoder, training auxiliary models, or learning concept-specific adapters.
\end{itemize}

\label{sec:related_works}
\section{Related Work}

\paragraph{Image editing and continuous control in diffusion models.}
Text-to-image diffusion and flow-matching models have enabled high-quality visual generation. However, precise editing remains challenging, as targeted changes must not interfere with unrelated content. 
A large body of work addresses this problem through prompt-based editing, inversion, attention control, feature injection, or instruction-tuned editing models. Prompt-to-Prompt~\citep{hertz2022prompt} controls cross-attention maps to preserve spatial layout under textual edits, while Null-text Inversion~\citep{mokady2022null} improves real-image editing by optimizing unconditional embeddings. Other methods manipulate internal diffusion features or attention maps to preserve structure during translation and editing~\citep{tumanyan2023plug, cao2023masactrl, brack2023Sega}. Instruction-based systems such as InstructPix2Pix~\citep{brooks2022instructpix2pix}, FLUX.1 Kontext~\citep{labs2025flux1kontextflowmatching}, and Qwen-Image~\citep{wu2025qwenimagetechnicalreport} provide a more direct natural-language editing interface, but the strength of an edit is often only indirectly controlled through the prompt.

Several works therefore introduce continuous or slider-like control. Concept Sliders~\citep{gandikota2023sliders} train concept-specific LoRA adapters whose scale controls the edit strength, while SliderSpace~\citep{gandikota2025sliderspace} discovers multiple controllable directions in diffusion models. Prompt Sliders~\citep{eccvsridhar24} instead use learned text embeddings, and recent instruction-based methods such as Kontinuous Kontext~\citep{parihar2025kontinuouskontext} and SliderEdit~\citep{zarei2025slideredit} expose continuous edit-strength controls for image editing models. Closest to our setting, FluxSpace~\citep{dalva2024fluxspace} edits rectified-flow transformers by manipulating internal transformer representations and supports both fine-grained and coarse edits. In contrast to these approaches, we do not train per-concept adapters, optimize edit-specific modules, or derive directions from a single edit condition. Instead, we study whether visual concept directions can be extracted directly from intermediate Diffusion Transformer activations using only a few positive and negative examples.

\paragraph{Representation-based concept steering for image generation.} A complementary line of work performs concept steering through learned or structured representation spaces~\citep{wang2022latent, 10645735, wang2025lia, wang2023leo} rather than directly extracting directions from DiT image activations. These methods are motivated by the observation that text prompts provide a natural interface for generation, but discrete prompts provide limited control over continuous edits and spatial localization. Prior work has identified semantic directions in token-level text embeddings for subject-specific control~\citep{baumann2025attributecontrol} or trained sparse autoencoders over text embeddings to expose disentangled edit directions~\citep{kamenetsky2025saedittokenlevelcontrolcontinuous}. Related work has also explored sparse autoencoders on U-Net activations and briefly tried on DiT, using learned sparse features to interpret or steer internal diffusion representations~\citep{surkov2024one, kim2025textitreveliointerpretingleveragingsemantic}. In a related activation-steering direction, CASteer~\citep{gaintseva2025casteer} computes concept-specific directions from averaged diffusion activations and applies them at inference time to cross-attention representations for concept erasure. While these methods demonstrate the value of internal diffusion representations, they either require training an auxiliary representation model or focus on suppressing unwanted concepts, rather than few-shot extraction of directions for controllable global and localized steering. Concurrent to our work, \citet{Ohib_2026_CVPR} extract and inject layer- and timestep-specific concept directions from contrastive prompt pairs in the residual stream of PixArt diffusion transformers.

Our work differs in where the concept direction is estimated. Rather than steering the text encoder or modifying token embeddings, we extract directions from image-side activations of the Diffusion Transformer itself. This distinction is important for visual concepts whose expression is tied to spatial image tokens rather than only to text semantics. We further show that the spatial support of the estimator matters: global difference-of-means is appropriate for concepts expressed throughout the image, while localized concepts such as object colors, textures, materials, or facial attributes require masked difference-of-means over the region where the concept appears. Thus, our method can be viewed as bringing the simplicity of contrastive direction extraction into the visual activation space, while explicitly matching the extraction procedure to the spatial structure of the concept.

\section{Method}
\label{sec:method}

\vspace{-0.2cm}

We propose a training-free method for extracting reusable visual concept directions from a frozen Diffusion Transformer. We first define the activation-space setup in Section~\ref{sec:method_setup}. We then introduce masked difference-of-means extraction in Section~\ref{sec:spatially_matched_extraction}.

\begin{figure}[t]
    \centering
    \includegraphics[width=0.9\textwidth]{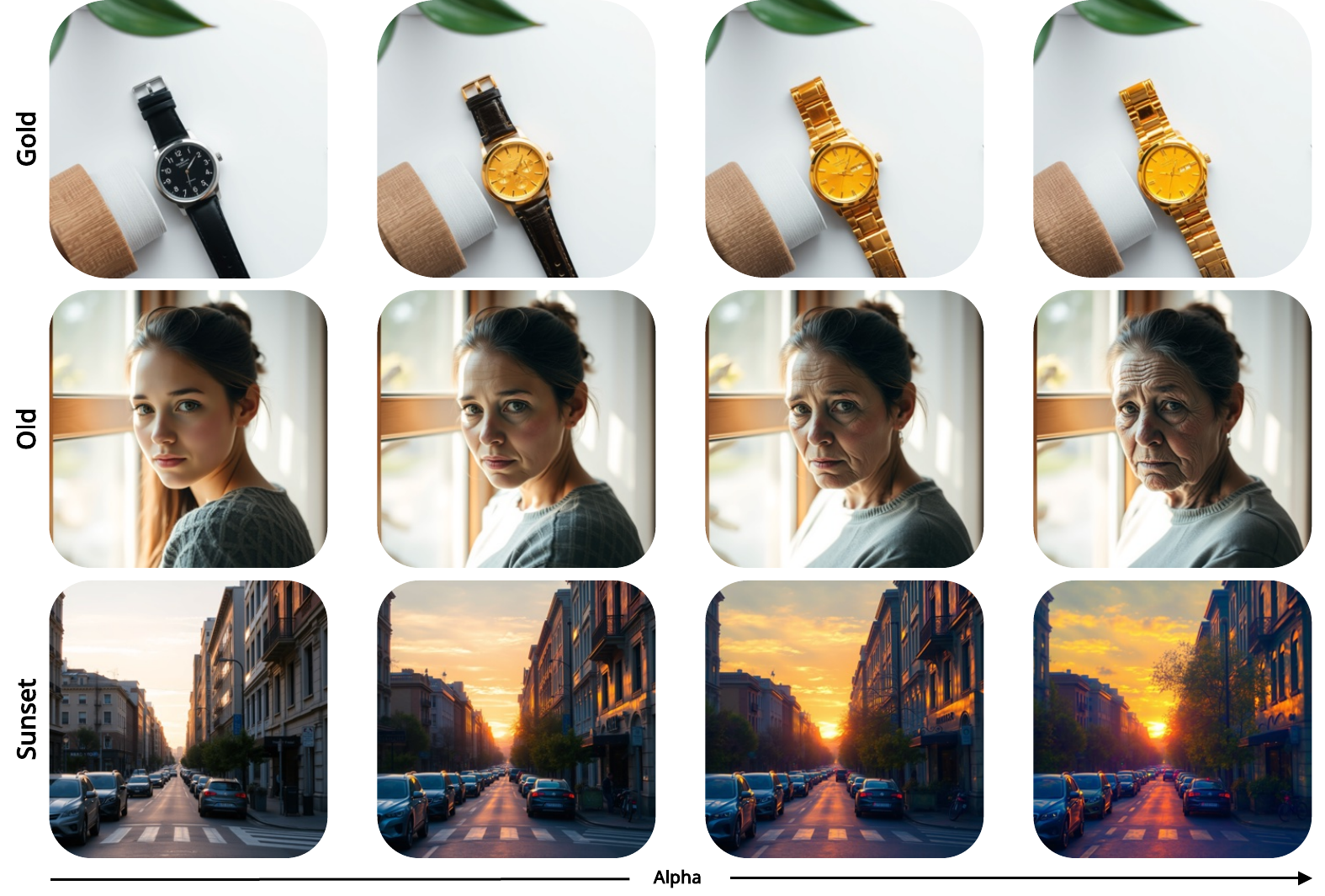}
    \caption{
    \textbf{Qualitative examples of concept steering.}
    We show representative examples of our method across different concept families. Each row corresponds to a held-out prompt and visualizes the effect of increasing the steering strength \(\alpha\). The examples illustrate that extracted concept directions can produce smooth steering while preserving the main content of the image.
    }
    \label{fig:steering_examples_placeholder}
    \vspace{-0.4cm}
\end{figure}

For localized concepts, we describe how image-space masks are converted into Diffusion Transformer token masks in Section~\ref{sec:token_selection_masks}. Finally, we explain how the extracted directions are used for inference-time steering in Sections~\ref{sec:inference_time_steering}. The central design choice is the spatial support used for estimation: global concepts are extracted from all image tokens, while localized concepts are extracted only from the tokens corresponding to the region where the concept appears.

\vspace{-0.2cm}

\subsection{Setup}
\label{sec:method_setup}

Let \(G_\theta\) denote a frozen text-to-image Diffusion Transformer (DiT). For a concept \(c\), we assume a small set of \(K\) paired prompts \(\mathcal{D}_c = \{(p_i^-, p_i^+)\}_{i=1}^{K}\), where \(p_i^+\) contains the target concept and \(p_i^-\) is a matched prompt in which the concept is absent or replaced by a neutral alternative. For example, a pair may differ by the presence of a color, texture, material, expression, or global style.

During generation, we record the image-token update produced by transformer
block $\ell$ at denoising step $t$. Let
$X^{\mathrm{in}}_{\ell,t}(p), X^{\mathrm{out}}_{\ell,t}(p)
\in \mathbb{R}^{N \times d}$ denote the image-token representations before
and after the block, respectively. We define the block update as
\begin{equation}
H_{\ell,t}(p)
=
X^{\mathrm{out}}_{\ell,t}(p)
-
X^{\mathrm{in}}_{\ell,t}(p),
\end{equation}
where $N$ is the number of image tokens and $d$ is the residual dimension.
Thus, concept directions are extracted from the residual update contributed
by the selected block, rather than from its absolute hidden representation.
In practice, we extract these updates from an intermediate transformer block,
where visual attributes are strongly represented. The model parameters are
never updated.

\subsection{Masked Difference-of-Means}
\label{sec:spatially_matched_extraction}

Our goal is to estimate a direction \(v_c \in \mathbb{R}^d\) corresponding to the visual concept \(c\). A natural estimator is difference-of-means: average the residual representation of positive examples and subtract the average residual representation of negative examples. However, the appropriate averaging region depends on the spatial structure of the concept.

For globally distributed concepts, such as lighting, style, atmosphere, or scene-level appearance, we average over all image tokens. For localized concepts, such as object colors, textures, materials, or facial attributes, we first identify the image region where the concept is instantiated and restrict the estimate to the corresponding image tokens. The construction of these token masks is described in Section~\ref{sec:token_selection_masks}.

We write this using a masked pooling operator
\begin{equation}
\mu(H, M) =
\frac{\sum_{j=1}^{N} M_j H_j}{\sum_{j=1}^{N} M_j + \epsilon},
\end{equation}

where \(H_j \in \mathbb{R}^d\) is the residual activation at image token \(j\), \(M \in [0,1]^N\) is a mask over Diffusion Transformer image tokens, and \(\epsilon\) is a small constant for numerical stability.

The concept direction is estimated by averaging the masked positive-negative differences across prompt pairs
\begin{equation}
v_c =
\frac{1}{K}
\sum_{i=1}^{K}
\left[
\mu(H_{\ell,t}(p_i^+), M_i^+)
-
\mu(H_{\ell,t}(p_i^-), M_i^-)
\right].
\end{equation}

This formulation unifies global and local extraction. For global concepts, we set \(M_i^+ = M_i^- = \mathbf{1}\), recovering standard difference-of-means over all image tokens. For local concepts, \(M_i^+\) and \(M_i^-\) select only the tokens corresponding to the concept region, reducing dilution from unrelated tokens. 

\subsection{Token Selection from Spatial Masks}
\label{sec:token_selection_masks}

For localized concepts, the mask is initially obtained in image space. We explored using cross-attention maps and ConceptAttention~\citep{helbling2025conceptattentiondiffusiontransformerslearn} to identify where concepts are expressed in the image, but found that these signals often do not segment the relevant regions reliably for attributes such as face regions or colors. We therefore use the SAM 3 segmentation model~\citep{carion2025sam3segmentconcepts} with a concept-specific text prompt to identify the relevant region, producing a pixel-space mask \(A_i \in [0,1]^{H \times W}\). This mask is then resized to the latent image-token resolution of the DiT, yielding a token mask \(M_i \in [0,1]^N\). Thus, masking is applied at the level of DiT image tokens rather than raw pixels.

Each token is selected according to the downsampled segmentation mask on the DiT token grid. This token selection step is crucial for localized concepts. If the concept occupies only a small part of the image, global pooling averages its activation change together with unrelated background, pose, and layout tokens. By projecting the spatial mask onto the DiT token grid, the masked difference-of-means estimator in Section~\ref{sec:spatially_matched_extraction} computes the direction from tokens that visually support the concept. 

\subsection{Inference-Time Steering}
\label{sec:inference_time_steering}

Once extracted, the concept direction \(v_c\) is used without normalization at inference time on new prompts. Let \(\mathcal{L}\) denote the set of transformer blocks where steering is applied, and let \(\mathcal{T}\) denote the denoising steps during which the intervention is active. At each selected block and step, we modify the image-token residuals as
\begin{equation}
H'_{\ell,t,j}
=
H_{\ell,t,j}
+
\alpha \, M^{\mathrm{steer}}_j \, v_c,
\qquad
\ell \in \mathcal{L}, \; t \in \mathcal{T}.
\end{equation}
Here, \(\alpha\) controls the steering strength and \(M^{\mathrm{steer}}\) determines the spatial region for the intervention. For global steering, the direction is applied to all image tokens. For local steering, it is applied only to the selected region.

The extraction mask determines how the concept direction is estimated from positive and negative examples, as described in Sections~\ref{sec:spatially_matched_extraction} and~\ref{sec:token_selection_masks}. For local steering, the steering mask is obtained by segmenting the target region with SAM~3 in a first unsteered generation.

\begin{figure}[t]
    \centering
    \includegraphics[width=\linewidth]{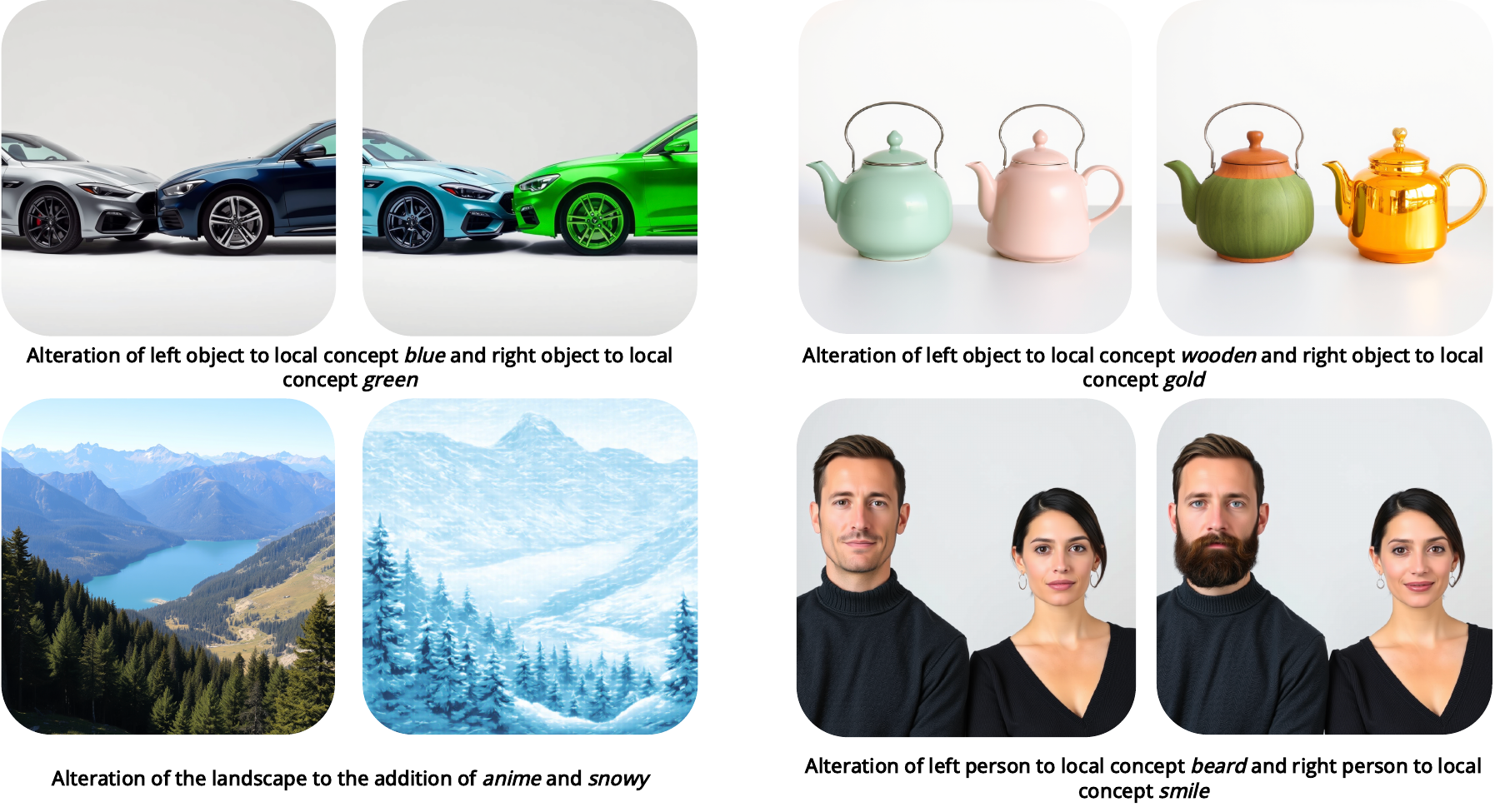}
    \caption{
    \textbf{Compositional and targeted steering.}
    Examples of different concept directions being applied to different regions in the same image. By selecting the spatial area to steer, our method enables targeted steering such as changing two different objects with two different colors or textures, as well as modifying different attributes of two persons. Notably, global concepts can be combined, as shown with \textit{snowy} $+$ \textit{anime} in a single inference steering pass.}
    \label{fig:compositional_edits_placeholder}
    \vspace{-0.5cm}
\end{figure}

\section{Experiments}
\label{sec:experiments}

\vspace{-0.2cm}

We evaluate whether few-shot activation-space directions can steer both global and localized visual concepts in a frozen Diffusion Transformer. Our experiments are designed to answer three questions: whether concept directions extracted from a small number of prompt pairs generalize to held-out prompts, whether masked extraction improves localized concept control and whether spatially masked steering enables multiple concepts to be composed in the same image. 

\vspace{-0.2cm}

\subsection{Implementation details}

We conduct our main experiments using FLUX.1-schnell, the 4-step distilled
variant, and use FLUX.1-dev for comparisons with SAEdit and FluxSpace to ensure
a comparable backbone. Unless otherwise stated, we extract concept directions
at layer 18, the last dual-stream block of the FLUX architecture, and inject
the resulting directions from layer 18 through layer 56, the final transformer
block, from the first, highest-noise step to the last. Section~\ref{subsec:ablation} supports these design choices. 
Moreover, preliminary
results with FLUX.2 in Appendix~\appref{app:flux2_real_image} suggest that the approach generalizes beyond FLUX.1 to newer reference-image editing models.


\subsection{Qualitative Results}

We show that our method is qualitatively effective and that simple difference-of-means steering with only 5 prompt pairs produces smooth and continuous steering trajectories. Figure~\ref{fig:steering_examples_placeholder} presents representative steering examples across different $\alpha$ strengths, and Figure~\ref{fig:compositional_edits_placeholder} shows that different steering vectors can be applied within the same image to control different regions, for example making one car blue and another green, or making one person smile while increasing the beard of another. Importantly, this compositional control does not require training on large collections of edit pairs. As these examples illustrate, our method supports both local and global steering. Additional examples, including weather changes, object and face steering, as well as real-image editing with FLUX.2, are provided in Appendices~\appref{app:more_qual} and~\appref{app:flux2_real_image}.

\begin{table}[t]
    \centering
    \caption{
    \textbf{Quantitative comparison across methods.}
    Higher $\Delta$VQA indicates stronger agreement with the requested edit, while lower LPIPS indicates better preservation of the original image.
    Methods are grouped by class.
    \editingmethod{Editing models} are trained image-editing models conditioned on the reference image and serve as an upper reference.
    \steeringmethod{Training-free} baselines use the FLUX.1-dev backbone and are directly comparable to our Dev variant.
    We report our method on FLUX.1-dev with 5 positive/negative prompt pairs, and on FLUX.1-schnell with 5, 10, and 20 pairs.
    }
    \label{table_main_res}
    \vspace{0.5em}
    \scriptsize
    \resizebox{\linewidth}{!}{
    \begin{tabular}{lcccccccc}
        \toprule
        & \multicolumn{2}{c}{Editing models}
        & \multicolumn{2}{c}{Training-free}
        & \multicolumn{4}{c}{Ours} \\
        \cmidrule(lr){6-9}
        & \multicolumn{2}{c}{\textit{(reference-conditioned)}}
        & \multicolumn{2}{c}{\textit{(concept steering)}}
        & Dev
        & \multicolumn{3}{c}{Schnell} \\
        \cmidrule(lr){2-3}
        \cmidrule(lr){4-5}
        \cmidrule(lr){6-6}
        \cmidrule(lr){7-9}
        Metric
        & \editingmethod{SliderEdit}
        & \editingmethod{K. Kontext}
        & \steeringmethod{SAEdit}
        & \steeringmethod{FluxSpace}
        & 5 pairs
        & 5 pairs
        & 10 pairs
        & 20 pairs \\
        \midrule
        $\Delta$VQA $\uparrow$
        & 0.35 & 0.36
        & 0.19 & 0.17
        & 0.29
        & 0.32 & 0.32 & 0.31 \\
        LPIPS $\downarrow$
        & 0.44 & 0.32
        & 0.28 & 0.21
        & 0.34
        & 0.29 & 0.29 & 0.29 \\
        \bottomrule
    \end{tabular}
    }
\end{table}

\subsection{Quantitative Results}

We quantitatively evaluate our method against both trained and
training-free baselines: SliderEdit~\citep{zarei2025slideredit}, Kontinuous Kontext~\citep{parihar2025kontinuouskontext}, SAEdit~\citep{kamenetsky2025saedittokenlevelcontrolcontinuous}, 
\begin{wrapfigure}{r}{0.48\columnwidth}
    \vspace{-0.4cm}
    \centering
    \includegraphics[width=\linewidth]{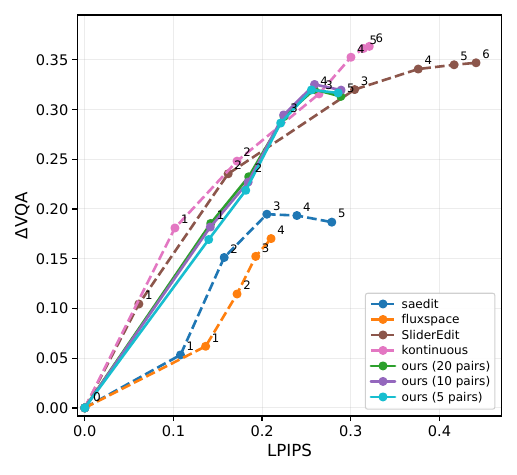}
    \caption{
    \textbf{Edit success versus content preservation.}
    We compare methods using \(\Delta\)VQA to measure whether the requested edit is achieved and LPIPS to measure visual drift from the original image. Better methods appear toward the upper-left, indicating stronger edit success with less content drift. We note that \(\Delta\)VQA slightly decreases at steering level 5, as some concepts appear at lower strengths and stronger interventions can lead to oversteering.
    }
    \label{fig_results_vqavslpips}
    \vspace{-0.3cm}
\end{wrapfigure}
and FluxSpace~\citep{dalva2024fluxspace}. We compare these methods in terms of their ability to provide continuous control across different steering strengths, while simultaneously achieving the requested edit and preserving the overall structure of the original image. For evaluation, we use category-specific steering strengths to account for differences in concept sensitivity. Face concepts are evaluated with \(\alpha \in \{0, 1, 1.5, 2, 2.5, 3\}\), object concepts with \(\alpha \in \{0, 2, 2.5, 3, 3.5, 4\}\), and global concepts with \(\alpha \in \{0, 2, 3, 4, 5, 6\}\). We use smaller strengths for face and object concepts because these edits typically appear earlier, while global concepts often require stronger interventions to produce visible scene-level changes. For all baselines, we use the editing procedures provided by the authors. For each of the 23 concepts, we use five images and three random seeds, resulting in a total of 2,070 generated images across the six steering strengths. No manual tuning per concept is applied and the $\alpha$ values presented are used for all concepts of the corresponding categories. In Table~\ref{table_main_res}, we report $\Delta$VQA and LPIPS at the strongest evaluated edit level of each method. In addition to the metrics reported in the main paper, we provide complementary preservation metrics, including DINO similarity and face similarity, in Appendix~\appref{app:additional_metrics}.

\subsubsection{Evaluation Suite}

We evaluate edits using two concept families: global and local concepts. \textbf{Global concepts} correspond to scene-level or style-level edits, including \textit{rainy}, \textit{sunset}, \textit{snowy}, \textit{cartoon}, and \textit{anime}. For these concepts, we use global difference-of-means. \textbf{Local concepts} correspond to spatially localized attributes and are divided into object and face concepts. Object concepts include attributes such as \textit{red}, \textit{blue}, \textit{green}, \textit{gold}, \textit{yellow}, \textit{metallic}, \textit{wooden}, \textit{furry}, \textit{shiny}, \textit{matte}, \textit{striped}, and \textit{dotted}. Face concepts include facial attributes and expressions such as \textit{smile}, \textit{laugh}, \textit{old}, \textit{beard}, \textit{glasses}, and \textit{lipstick}. For all local concepts, we use our proposed \textit{masked difference-of-means} for concept extraction. We specify the hyperparameters used for evaluation of these methods in Appendix~\appref{app:baseline_eval_details}. Evaluation prompts are distinct from extraction prompts, although some share scene categories.

\subsubsection{Metrics}

To evaluate the compared methods, we measure both edit success and content preservation across different strengths. A good method should increase the target attribute while avoiding excessive drift from the original image. We therefore use \(\Delta\)VQA~\citep{lin2024evaluating} to measure edit success and LPIPS~\citep{zhang2018perceptual} to measure visual drift from the unedited image. Higher \(\Delta\)VQA indicates stronger agreement with the requested edit, while lower LPIPS indicates better preservation of the original image structure. More details about the metrics are available in Appendix~\appref{app:metric_details}.

Figure~\ref{fig_results_vqavslpips} and Table~\ref{table_main_res} show that our method approaches the results of trained editing methods while using only five concept pairs and no additional parameters or curated dataset and achieves substantially stronger edits than SAEdit and FluxSpace, whose low LPIPS mainly reflects limited edits, as shown by their low $\Delta$VQA. SliderEdit and Kontinuous Kontext are instead trained editing models conditioned on the reference image. All methods share the FLUX.1 architecture, since FLUX.1 Kontext builds on FLUX.1-dev with added image conditioning.  We also evaluate our method with different numbers of positive and negative prompt pairs for concept extraction, using 5, 10, and 20 pairs. We observe only marginal differences across these settings, suggesting that five pairs is sufficient to capture the main concept direction in activation space. 

\vspace{-0.25cm}

\subsection{Ablation of the Method}
\label{subsec:ablation}

To further validate our method, we ablate the main design choices involved in concept extraction and steering. We study the effect of the extraction layer, the denoising timesteps used to compute the concept direction, the layer range over which the direction is injected at inference time and also the effect of masked vs. global difference-of-means for local concepts. These ablations motivate our default configuration and identify which components are most important for effective steering. 
Finally, we discuss failure cases in Appendix~\appref{app:failure_cases}. More ablations are available in the appendix.

\begin{figure}[t]
    \centering
    \begin{minipage}[t]{0.48\linewidth}
        \vspace{0pt}
        \centering
        \includegraphics[width=\linewidth]{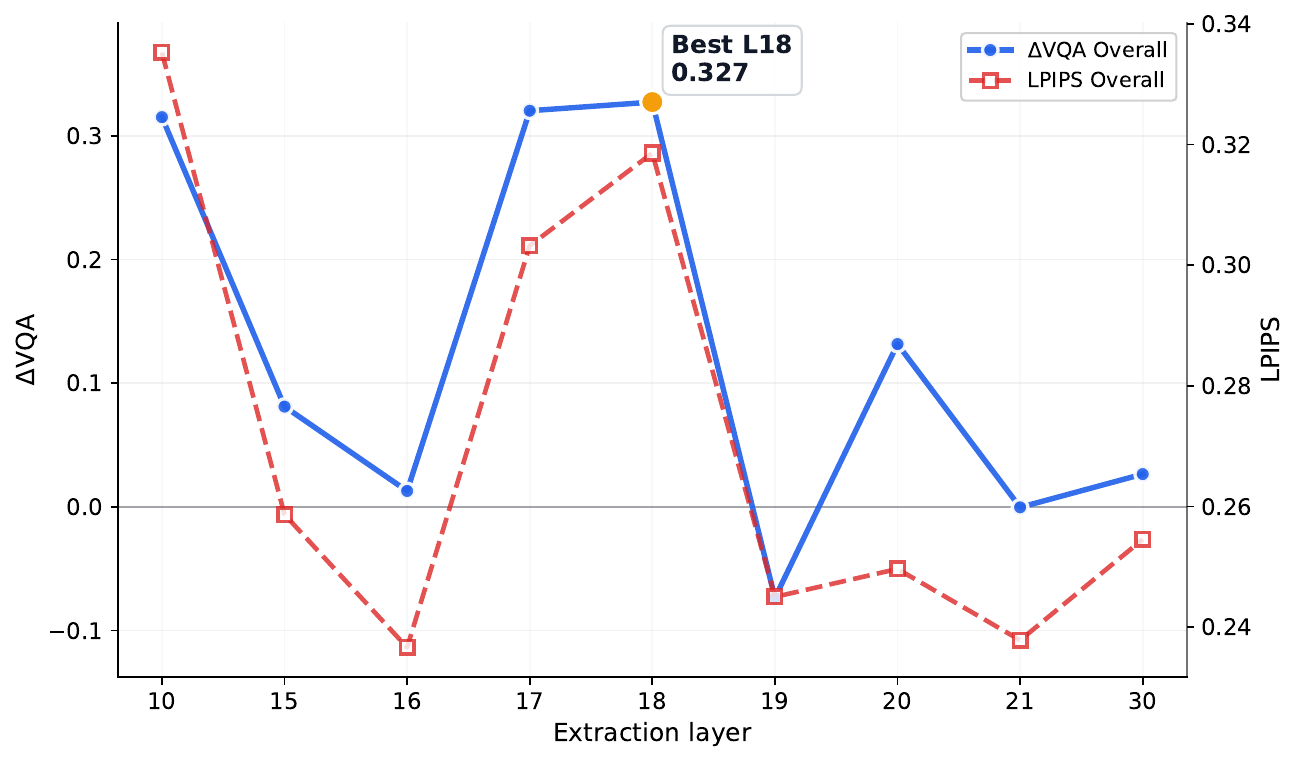}
        \captionof{figure}{
        \textbf{Extraction layer ablation.}
        We sweep the layer used to extract the concept direction while keeping the injection range fixed. Layer 18 provides the strongest trade-off between edit success and content preservation.
        }
        \label{fig:ablation_extraction_layer}
    \end{minipage}
    \hfill
    \begin{minipage}[t]{0.48\linewidth}
        \vspace{0pt}
        \centering
        \includegraphics[width=\linewidth]{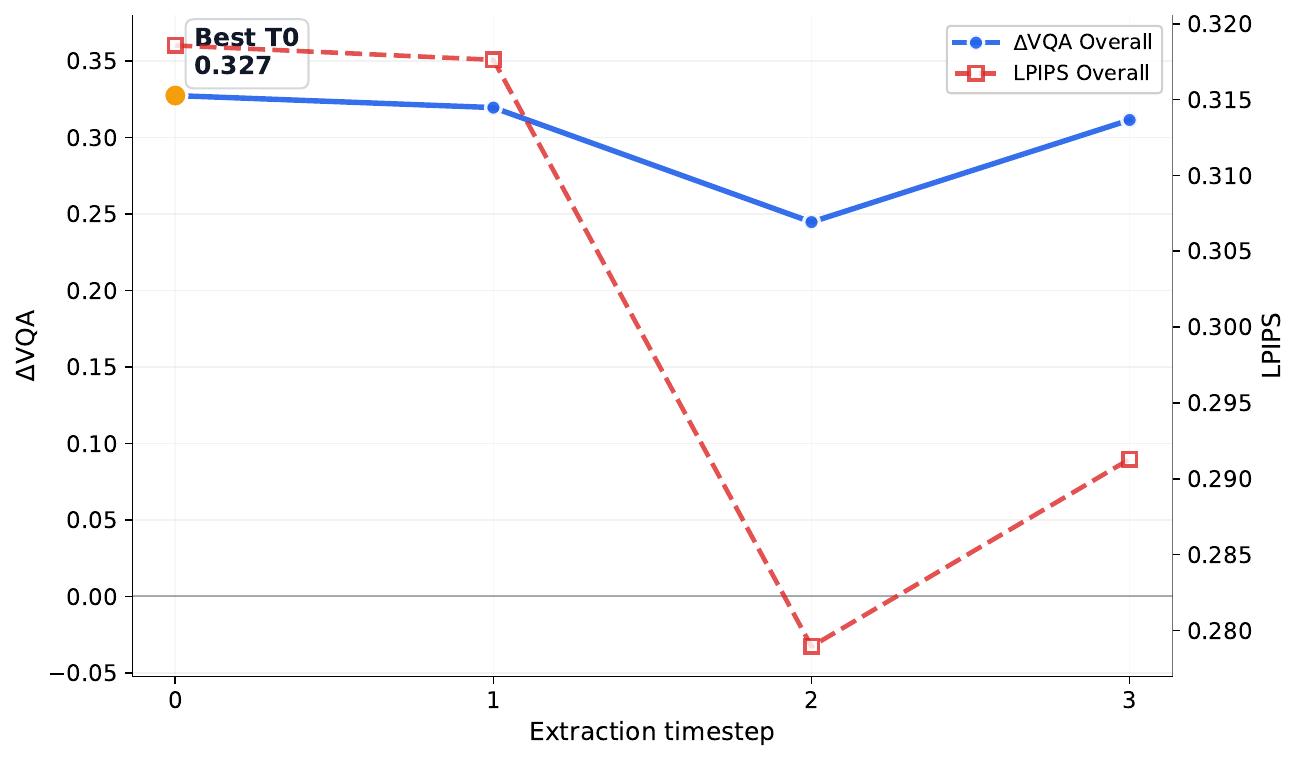}
        \captionof{figure}{
        \textbf{Extraction timestep ablation.}
        We compare concept directions extracted from individual denoising timesteps. The first denoising step provides the strongest edits.
        }
        \label{fig:ablation_extraction_timestep}
    \end{minipage}
\end{figure}

\subsubsection{Extraction Layer Ablation}

We identify the layer that yields the most effective concept directions, using one representative concept per category: \textit{snowy} (global), \textit{blue} (object), and \textit{beard} (face). As shown in Figure~\ref{fig:ablation_extraction_layer}, layer 18 achieves the best trade-off, with the highest $\Delta$VQA. Layer 17 performs slightly worse and qualitatively introduces over-exposure and artifacts for global concepts, alters object structure, and yields weaker face attributes at the same strength. Layer 10 also reaches a high $\Delta$VQA but with higher LPIPS, showing the same failure modes with more severe distortion. The remaining layers often fail to introduce the concept at all, suggesting the concept signal is not equally accessible throughout the network. We therefore use layer 18 by default; samples for each layer are in Appendix~\appref{app:qualitative_ablation}.

\subsubsection{Extraction Timestep Ablation}

We also ablate which denoising timestep should be used to extract the concept direction. We perform this ablation using the same representative concepts as in the layer study. As shown in Figure~\ref{fig:ablation_extraction_timestep}, timestep 0 provides the best qualitative behavior across categories and yields the most reliable concept directions.

Timestep 1 also produces visible changes but introduces artifacts: blur or object distortion for objects, weak or absent attributes for faces, and slight noise for global concepts. Later timesteps are less reliable: timestep 2 yields much weaker edits, and timestep 3, despite a high $\Delta$VQA, misplaces or distorts object edits, keeps face attributes weak, and only partially changes global scenes. Based on these results, we use timestep 0 as our default extraction timestep. Additional qualitative samples for each timestep are provided in Appendix~\appref{app:qualitative_ablation}.

\begin{table*}[t]
    \centering

    \begin{minipage}[t]{0.47\textwidth}
        \centering
        \small
        \captionof{table}{
        \textbf{Injection range ablation.}
        We compare several injection ranges while keeping the extraction configuration fixed. Blocks 18--56 gives the best overall trade-off.
        }
        \label{tab:ablation_injection_range}
        \vspace{0.3em}

        \begin{tabular}{lcc}
            \toprule
            Blocks
            & LPIPS $\downarrow$
            & $\Delta$VQA $\uparrow$ \\
            \midrule
            All    & 0.656 & -0.034 \\
            10--18 & 0.237 & 0.131 \\
            18--36 & 0.296 & 0.278 \\
            42--56 & {0.077} & 0.051 \\
            18--56 & 0.286 & {0.317} \\
            \bottomrule
        \end{tabular}
    \end{minipage}
    \hfill
    \begin{minipage}[t]{0.50\textwidth}
        \centering
        \small
        \captionof{table}{
        \textbf{Spatial masking ablation.}
        We compare concept directions extracted with and without spatial
        masking. Masking improves both editing and preservation, especially for face concepts.
        }
        \label{tab:ablation_masked_dom}
        \vspace{0.3em}

        \setlength{\tabcolsep}{5pt}
        \begin{tabular}{llcc}
            \toprule
            Concept
            & Extraction
            & LPIPS $\downarrow$
            & $\Delta$VQA $\uparrow$ \\
            \midrule
            Face
            & Masked DoM
            & \textbf{0.189}
            & \textbf{0.185} \\
            Face
            & Global DoM
            & 0.267
            & 0.171 \\
            Objects
            & Masked DoM
            & \textbf{0.164}
            & \textbf{0.287} \\
            Objects
            & Global DoM
            & 0.183
            & 0.277 \\
            \bottomrule
        \end{tabular}
    \end{minipage}

    \vspace{-0.2cm}
\end{table*}

\subsubsection{Injection Range Ablation}

We study the range of transformer blocks over which the concept
direction is injected at inference time. We keep the extraction layer and
timestep fixed to their default values and vary only the injection range.
Unlike the previous ablations, we evaluate all concepts and report the
average performance across face, global, and object.

As shown in Table~\ref{tab:ablation_injection_range}, injecting over blocks
18--56 gives the best overall trade-off, achieving the highest average
$\Delta$VQA while keeping LPIPS moderate. Using only later blocks
(42--56) better preserves the input but produces much weaker edits, while
injecting into all blocks strongly degrades both preservation and steering. Although the direction is extracted from a dual-stream block, it transfers
to the subsequent single-stream blocks, since all blocks update the same
image-token residual stream. We therefore use blocks 18--56 as our default injection range.

\subsubsection{Spatial Masking Ablation}

We ablate the use of spatial masking when computing the difference-of-means
concept direction. As shown in Table~\ref{tab:ablation_masked_dom}, masking improves steering
performance and preservation for both categories, with the largest gain in
preservation for face concepts. For face concepts, LPIPS decreases from 0.267
to 0.189 while $\Delta$VQA increases from 0.171 to 0.185. For object concepts,
$\Delta$VQA increases from 0.277 to 0.287, while LPIPS decreases from 0.183
to 0.164.
These results support using spatial masking during concept extraction. Appendix~\appref{app:lpips_outside} further shows that our edits remain largely
within the target region.

\section{Conclusions}
\label{sec:conclusion}

We introduced a training-free framework for extracting visual concept directions from Diffusion Transformer activations using few positive and negative examples. Effective steering requires matching extraction to the spatial structure of the concept: global difference-of-means is effective for image-wide concepts, while masked difference-of-means better captures local concepts otherwise diluted by global averaging. The directions support continuous global, local, and compositional steering without modifying model parameters, providing lightweight, fine-grained visual control in Diffusion Transformers.

Despite effective global and localized steering, our method exhibits some limitations. High steering strengths may slightly alter object shape, pose, or boundaries. Global steering can emphasize local attributes over coherent scene transformations, \textit{e.g.}, snowy may whiten objects without creating a full snowy scene, depending on the steering region’s compatibility with the concept. Facial concepts may also introduce correlated attributes (\textit{e.g.}, lipstick increasing smiling). Future work will focus on reducing spatial drift, improving concept disentanglement, and extending to video generation.

\begin{ack}
This work has been supported by the French government, through the 3IA Cote d’Azur Investments in the project managed by the National Research Agency (ANR) with the reference number ANR-23-IACL-0001. This project was provided with computer and storage resources by GENCI at IDRIS, thanks to the grant 2025-AD011015982R1 on the supercomputer Jean Zay’s H100 partition. The authors declare no competing interests.
\end{ack}

\bibliographystyle{plainnat}
\bibliography{neurips_2025}


\clearpage
\appendix

\renewcommand{\topfraction}{0.9}
\renewcommand{\bottomfraction}{0.8}
\renewcommand{\textfraction}{0.07}
\renewcommand{\floatpagefraction}{0.85}

\begin{center}
    {\Large\bfseries Appendix}
\end{center}
\vspace{0.5em}
\noindent This appendix provides additional implementation details, evaluation settings, ablations, and qualitative results. It is organized as follows.
\vspace{0.5em}

\etocsettocstyle{\noindent\rule{\linewidth}{0.4pt}\par\vspace{0.3em}}{\par\vspace{0.3em}\noindent\rule{\linewidth}{0.4pt}\par}
\etocsetstyle{section}
    {}
    {\addvspace{0.25em}}
    {\noindent\makebox[2em][l]{\textbf{\etocnumber}}\etocname\nobreak\leaders\hbox to 0.6em{\hss.\hss}\hfill\nobreak\makebox[1.5em][r]{\etocpage}\par}
    {}
\etoctoccontentsline{part}{Appendix}
\etocsetnexttocdepth{section}
\begingroup\hypersetup{linkcolor=black}
\localtableofcontents
\endgroup
\vspace{1em}

\section{Concept Extraction Prompts}
\label{app:concept_extraction_prompts}

This appendix describes the prompt pairs used to extract visual concept directions. We include two representative concepts for each concept family: global concepts, object concepts, and face concepts. Each pair consists of a positive prompt \(p_i^+\), which contains the target concept, and a matched negative prompt \(p_i^-\), where the target concept is removed or replaced by a neutral alternative.

For global concepts, no spatial localization is required: the extraction mask covers the full image. For localized concepts, we use SAM~3 to segment the target region and project the resulting mask to the DiT image-token grid. For object concepts, the target mask is the carrier object named in the prompt. For face concepts, the target mask is either the full face region or a tighter mouth region, depending on the attribute. We used an LLM, namely GPT 5.4 to generate the positive and negative pairs.

\subsection{Global Concepts}

\paragraph{Anime style.}
This concept captures a global rendering style. The extraction mask is the full image for all prompt pairs.

\begin{itemize}[leftmargin=*]
    \item Positive: ``A beach in daylight, rendered in anime style.'' \\
    Negative: ``A beach in daylight, as a normal photograph.''

    \item Positive: ``A mountain road in daylight, rendered in anime style.'' \\
    Negative: ``A mountain road in daylight, as a normal photograph.''

    \item Positive: ``An empty city street in daylight, rendered in anime style.'' \\
    Negative: ``An empty city street in daylight, as a normal photograph.''

    \item Positive: ``A forest path in daylight, rendered in anime style.'' \\
    Negative: ``A forest path in daylight, as a normal photograph.''

    \item Positive: ``A lake shore in daylight, rendered in anime style.'' \\
    Negative: ``A lake shore in daylight, as a normal photograph.''
\end{itemize}

\paragraph{Sunset lighting.}
This concept captures a global change in illumination and color temperature. The extraction mask is the full image for all prompt pairs.

\begin{itemize}[leftmargin=*]
    \item Positive: ``A beach in daylight, at sunset with warm orange light.'' \\
    Negative: ``A beach in daylight, under neutral daylight.''

    \item Positive: ``A mountain road in daylight, at sunset with warm orange light.'' \\
    Negative: ``A mountain road in daylight, under neutral daylight.''

    \item Positive: ``An empty city street in daylight, at sunset with warm orange light.'' \\
    Negative: ``An empty city street in daylight, under neutral daylight.''

    \item Positive: ``A forest path in daylight, at sunset with warm orange light.'' \\
    Negative: ``A forest path in daylight, under neutral daylight.''

    \item Positive: ``A lake shore in daylight, at sunset with warm orange light.'' \\
    Negative: ``A lake shore in daylight, under neutral daylight.''
\end{itemize}

\subsection{Object Concepts}

\paragraph{Green object color.}
This concept captures a localized object color. For each pair, the extraction mask is a SAM~3 mask of the carrier object, such as the car, bowl, chair, sneakers, or backpack.

\begin{itemize}[leftmargin=*]
    \item Target mask: car. \\
    Positive: ``A green car on a plain studio background.'' \\
    Negative: ``A gray car on a plain studio background.''

    \item Target mask: bowl. \\
    Positive: ``A green bowl on a plain tabletop.'' \\
    Negative: ``A gray bowl on a plain tabletop.''

    \item Target mask: chair. \\
    Positive: ``A green chair on a plain background.'' \\
    Negative: ``A gray chair on a plain background.''

    \item Target mask: sneakers. \\
    Positive: ``A pair of green sneakers on a plain studio background.'' \\
    Negative: ``A pair of gray sneakers on a plain studio background.''

    \item Target mask: backpack. \\
    Positive: ``A green backpack on a plain background.'' \\
    Negative: ``A gray backpack on a plain background.''
\end{itemize}

\paragraph{Gold object appearance.}
This concept captures both localized color and reflectivity. For each pair, the extraction mask is a SAM~3 mask of the carrier object. The negative prompt keeps the same object in gray to isolate the gold appearance.

\begin{itemize}[leftmargin=*]
    \item Target mask: car. \\
    Positive: ``A studio product photo of a gold car on a white background.'' \\
    Negative: ``A studio product photo of a gray car on a white background.''

    \item Target mask: chair. \\
    Positive: ``A centered gold chair on a plain light-gray backdrop.'' \\
    Negative: ``A centered gray chair on a plain light-gray backdrop.''

    \item Target mask: mug. \\
    Positive: ``A realistic photo of a gold mug on a wooden table, plain background.'' \\
    Negative: ``A realistic photo of a gray mug on a wooden table, plain background.''

    \item Target mask: backpack. \\
    Positive: ``A tabletop photo of a gold backpack with soft lighting, uncluttered background.'' \\
    Negative: ``A tabletop photo of a gray backpack with soft lighting, uncluttered background.''

    \item Target mask: sneaker. \\
    Positive: ``A studio product photo of a gold sneaker on a white background.'' \\
    Negative: ``A studio product photo of a gray sneaker on a white background.''
\end{itemize}

\subsection{Face Concepts}

\paragraph{Older facial appearance.}
This concept captures age-related facial attributes, such as wrinkles and visible age lines. The extraction mask is a SAM~3 mask of the face region.

\begin{itemize}[leftmargin=*]
    \item Target mask: face. \\
    Positive: ``A close-up portrait of a young woman with an elderly face and visible age lines, looking at the camera.'' \\
    Negative: ``A close-up portrait of a young woman with a youthful face and smooth skin, looking at the camera.''

    \item Target mask: face. \\
    Positive: ``A close-up portrait of a middle-aged woman with an elderly face and visible age lines, looking at the camera.'' \\
    Negative: ``A close-up portrait of a middle-aged woman with a youthful face and smooth skin, looking at the camera.''

    \item Target mask: face. \\
    Positive: ``A passport-style photo of a woman with an elderly face and visible age lines, against a plain background.'' \\
    Negative: ``A passport-style photo of a woman with a youthful face and smooth skin, against a plain background.''

    \item Target mask: face. \\
    Positive: ``A shoulder-up portrait of a woman with an elderly face and visible age lines, against a plain wall.'' \\
    Negative: ``A shoulder-up portrait of a woman with a youthful face and smooth skin, against a plain wall.''

    \item Target mask: face. \\
    Positive: ``A studio portrait of a woman with an elderly face and visible age lines, against a plain gray background.'' \\
    Negative: ``A studio portrait of a woman with a youthful face and smooth skin, against a plain gray background.''
\end{itemize}

\paragraph{Smile expression.}
This concept captures a localized facial expression. The extraction mask is a tight SAM~3 mask of the mouth region, which helps limit drift in unrelated facial features.

\begin{itemize}[leftmargin=*]
    \item Target mask: mouth. \\
    Positive: ``A close-up portrait of a young woman smiling warmly, looking at the camera.'' \\
    Negative: ``A close-up portrait of a young woman with a neutral expression, looking at the camera.''

    \item Target mask: mouth. \\
    Positive: ``A close-up portrait of a middle-aged woman smiling warmly, looking at the camera.'' \\
    Negative: ``A close-up portrait of a middle-aged woman with a neutral expression, looking at the camera.''

    \item Target mask: mouth. \\
    Positive: ``A passport-style photo of a woman smiling warmly, against a plain background.'' \\
    Negative: ``A passport-style photo of a woman with a neutral expression, against a plain background.''

    \item Target mask: mouth. \\
    Positive: ``A shoulder-up portrait of a woman smiling warmly, against a plain wall.'' \\
    Negative: ``A shoulder-up portrait of a woman with a neutral expression, against a plain wall.''

    \item Target mask: mouth. \\
    Positive: ``A studio portrait of a woman smiling warmly, against a plain gray background.'' \\
    Negative: ``A studio portrait of a woman with a neutral expression, against a plain gray background.''
\end{itemize}

\section{Evaluation Metric Details}
\label{app:metric_details}

This appendix provides additional details on the metrics used in our quantitative evaluation. Reliably evaluating generated content remains an open challenge, as also studied for talking head video generation \cite{Quignon_2026_CVPR}. We evaluate each steered image along two complementary axes: edit success and content preservation. Edit success is measured with \(\Delta\)VQA, while content preservation is measured with LPIPS.

\subsection{($\Delta$)VQA}

We use VQAScore~\citep{lin2024evaluating} to measure whether the generated image better matches the requested target concept after steering. In our implementation, VQAScore is computed using the \texttt{clip-flant5-xxl} model, the default model of the \texttt{t2v\_metrics} package.

For each evaluation sample, we construct a target text describing the desired edited result. For object concepts, the target text inserts the desired attribute before the relevant object. For example, if the original prompt contains a car and the concept is \textit{green}, the target text describes a green car. For face and global concepts, the target attribute is appended to the prompt, such as adding \textit{smiling}, \textit{old}, \textit{anime style}, or \textit{sunset lighting}.

We score both the original image and the steered image against the same target text. Let \(x_0\) denote the original image, \(x_\alpha\) the steered image at strength \(\alpha\), and \(q\) the target text. We define
\[
\Delta\mathrm{VQA}
=
\mathrm{VQAScore}(x_\alpha, q)
-
\mathrm{VQAScore}(x_0, q).
\]
This measures how much steering increases agreement with the requested concept relative to the unsteered image. Positive values indicate that the steered image better matches the target concept than the original image.

\subsection{LPIPS}

We use LPIPS~\citep{zhang2018perceptual} to measure perceptual drift from the original image. In our implementation, LPIPS is computed with the official \texttt{lpips} package using the AlexNet backbone, i.e., \texttt{net=alex}.

For each prompt and random seed, we compare every steered variant to its matching original image. Both images are loaded in RGB format and normalized to the \([-1,1]\) range before scoring. If \(x_0\) is the original image and \(x_\alpha\) is the steered variant, the reported score is
\[
\mathrm{LPIPS}(x_0, x_\alpha).
\]
Lower LPIPS indicates that the steered image remains perceptually closer to the original image, while higher LPIPS indicates larger visual changes. In our evaluation, LPIPS is therefore used as a measure of content preservation, complementary to \(\Delta\)VQA.

\subsection{Aggregation}

For both metrics, we compute scores for each prompt, concept, random seed, and steering strength. We then aggregate results by averaging across the corresponding evaluation set. In the main paper, we report the average $\Delta$VQA and LPIPS at each evaluated edit strength, and Table~\ref{table_main_res} reports the strongest evaluated edit level. A good method should obtain high \(\Delta\)VQA while keeping LPIPS low, indicating that it introduces the requested concept without unnecessarily changing unrelated image content.

\section{Baseline Evaluation Details}
\label{app:baseline_eval_details}

This appendix summarizes the main evaluation settings used for the compared baselines. We focus on the model variants and edit-strength values used in the experiments.

\subsection{SAEdit}

For SAEdit~\citep{kamenetsky2025saedittokenlevelcontrolcontinuous}, we use the released implementation with the released SAE checkpoint \texttt{Ronenk94/T5\_matryoshka\_sae} on \texttt{black-forest-labs/FLUX.1-dev}. For the all-method comparison, steering directions are computed from 100 sentence pairs per concept, using our evaluation-suite pairs together with templated extensions.

We evaluate the following edit-strength values:
\[
\alpha \in \{0,\;0.5,\;1.0,\;1.5,\;2.0,\;3.0\}.
\]

\subsection{FluxSpace}

For FluxSpace~\citep{dalva2024fluxspace}, we use the released training-free implementation with \texttt{black-forest-labs/FLUX.1-dev}. We evaluate the following edit levels:
\[
\lambda \in \{0,\;0.25,\;0.5,\;0.75,\;1.0\}.
\]
The edit level multiplies the configured steering scales, so \(\lambda=1.0\) corresponds to the strongest evaluated FluxSpace edit.

For local edits, used for object and face concepts, we use the local-edit configuration with edit timesteps from 3 to 30. For global and stylization concepts, we use the global-edit configuration with edit timesteps from 0 to 3.

\subsection{SliderEdit}

For SliderEdit~\citep{zarei2025slideredit}, we use the released FLUX Kontext GSTLoRA pipeline with \texttt{black-forest-labs/FLUX.1-Kontext-dev} and the checkpoint \texttt{example\_training\_gstlora\_iter500.safetensors}. SliderEdit is evaluated in the image-editing setting using precomputed original images as inputs.

We evaluate the following alpha values:
\[
\alpha \in \{-1.0,\;-0.6,\;-0.2,\;0.2,\;0.6,\;1.0\}.
\]
For this released GSTLoRA checkpoint, negative alpha strengthens the edit while positive alpha suppresses it. Therefore, the strongest evaluated SliderEdit setting is \(\alpha=-1.0\).

\subsection{Kontinuous Kontext}

For Kontinuous Kontext~\citep{parihar2025kontinuouskontext}, we use \texttt{black-forest-labs/FLUX.1-Kontext-dev} with the released LoRA weights and slider projector. The method is evaluated as an image-conditioned editing model using precomputed source images.

We set the number of edit steps to 6, producing the following edit levels:
\[
\lambda \in
\left\{
0,\;
\frac{1}{6},\;
\frac{2}{6},\;
\frac{3}{6},\;
\frac{4}{6},\;
\frac{5}{6},\;
1
\right\}.
\]

\section{Qualitative Ablation Samples}
\label{app:qualitative_ablation}

This appendix provides qualitative samples for the extraction-layer and extraction-timestep ablations. Each example keeps the evaluation prompt, random seed, steering strength, and inference-time injection range fixed, while varying the extraction setting used to obtain the concept direction. These examples complement the aggregate ablation plots in Section~\ref{subsec:ablation} by showing how different extraction choices affect the visual behavior of the same concept on the same prompt.

Figures~\ref{fig:qual_ablation_beard}, \ref{fig:qual_ablation_blue}, and~\ref{fig:qual_ablation_snowy} show three samples for a face concept (\textit{beard}), an object concept (\textit{blue}), and a global concept (\textit{snowy}), respectively. In each panel, the top row shows the extraction-timestep ablation and the bottom row shows the extraction-layer ablation.

\newcommand{\QualAblationPanel}[3]{%
    \begin{subfigure}[t]{\linewidth}
        \centering
        \includegraphics[width=0.62\linewidth]{figs/ablation/#1}\\[0.3em]
        \includegraphics[width=\linewidth]{figs/ablation/#2}
        \caption{Prompt: ``#3''}
    \end{subfigure}%
}

\begin{figure}[!htbp]
    \centering
    \QualAblationPanel
    {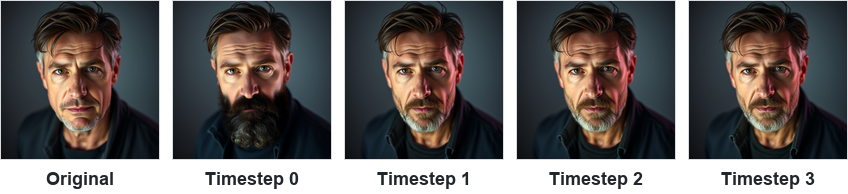}
    {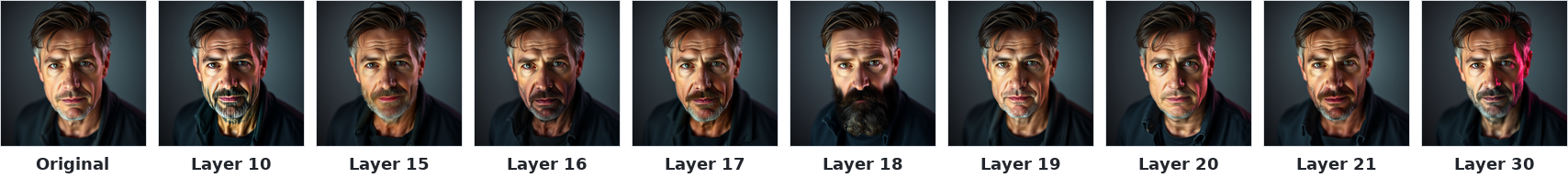}
    {A close-up portrait of a middle-aged man looking at the camera in studio light.}

    \vspace{0.8em}

    \QualAblationPanel
    {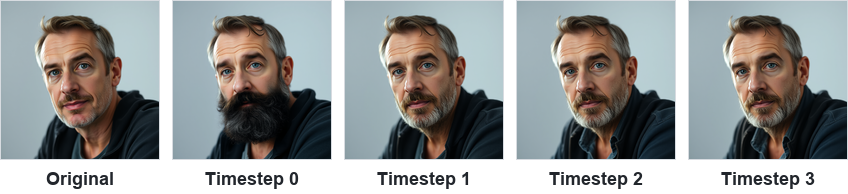}
    {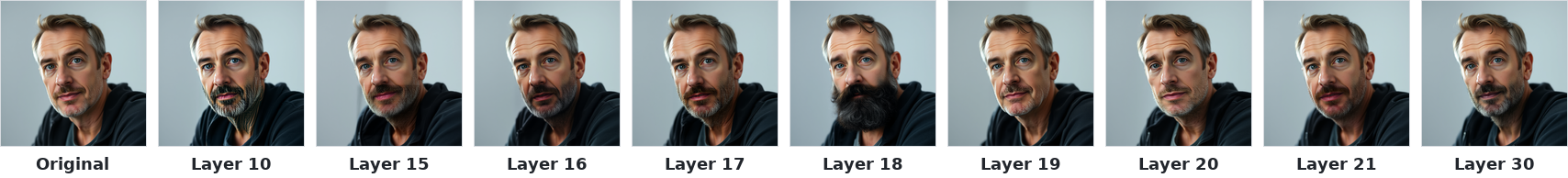}
    {A close-up portrait of a middle-aged man looking at the camera in studio light.}

    \vspace{0.8em}

    \QualAblationPanel
    {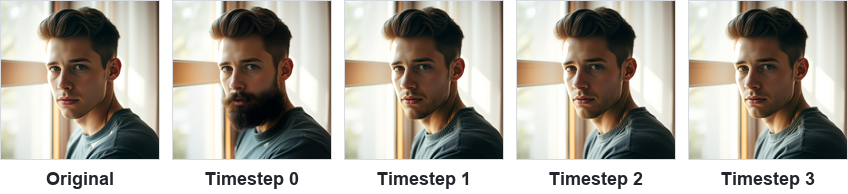}
    {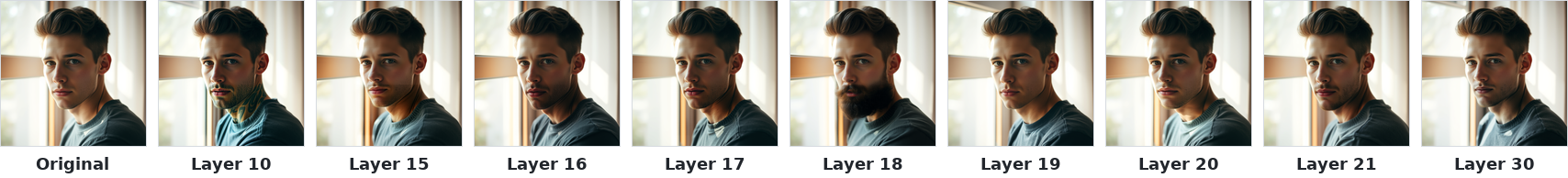}
    {A close-up portrait of a young man looking at the camera in soft window light.}

    \caption{
    \textbf{Qualitative ablation samples -- face concept: \textit{beard}.}
    Top row of each panel: extraction-timestep ablation. Bottom row: extraction-layer ablation.
    }
    \label{fig:qual_ablation_beard}
\end{figure}

\begin{figure}[!htbp]
    \centering
    \QualAblationPanel
    {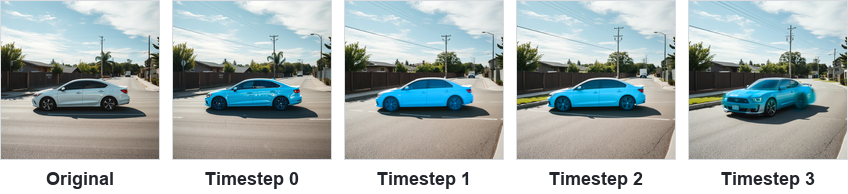}
    {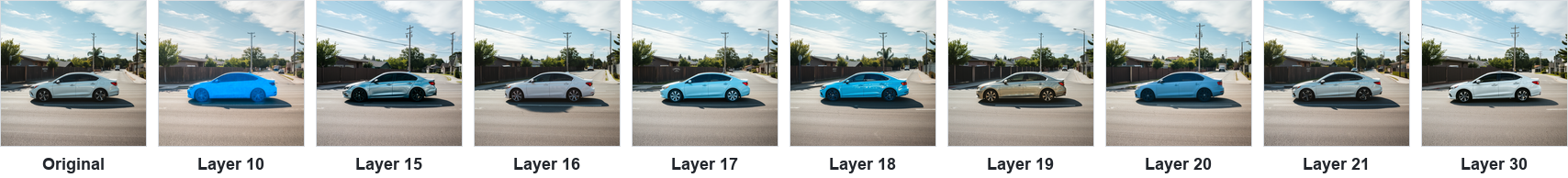}
    {A sedan parked on a quiet street.}

    \vspace{0.8em}

    \QualAblationPanel
    {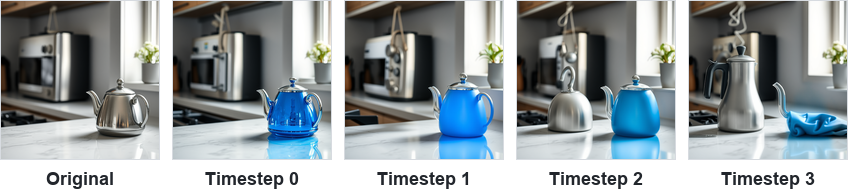}
    {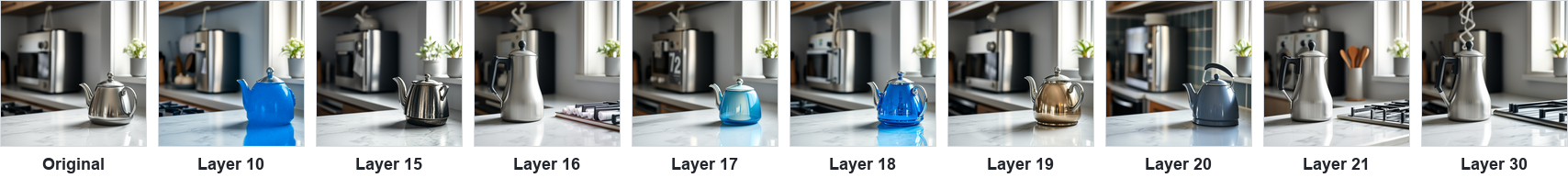}
    {A teapot on a kitchen counter.}

    \vspace{0.8em}

    \QualAblationPanel
    {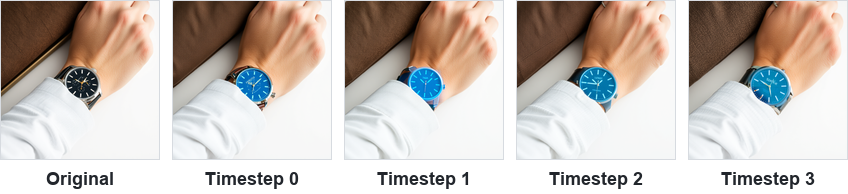}
    {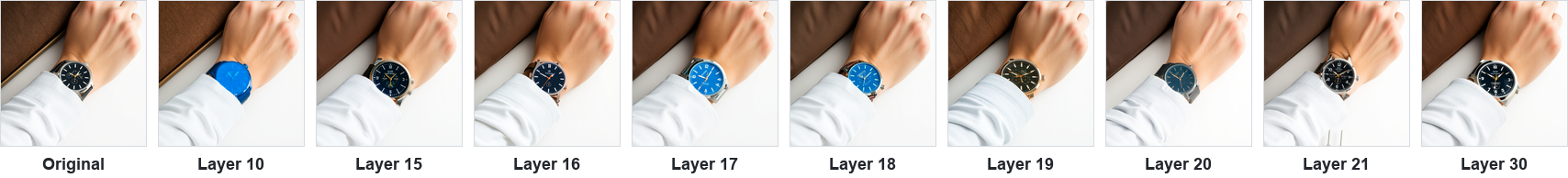}
    {A wristwatch on a white tabletop.}

    \caption{
    \textbf{Qualitative ablation samples -- object concept: \textit{blue}.}
    Top row of each panel: extraction-timestep ablation. Bottom row: extraction-layer ablation.
    }
    \label{fig:qual_ablation_blue}
\end{figure}

\begin{figure}[!htbp]
    \centering
    \QualAblationPanel
    {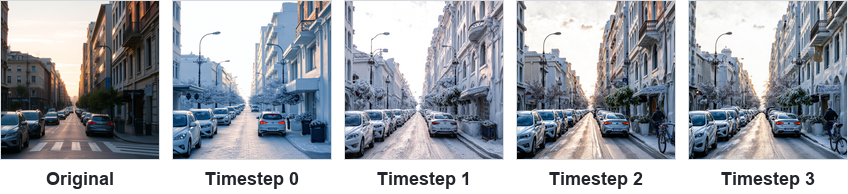}
    {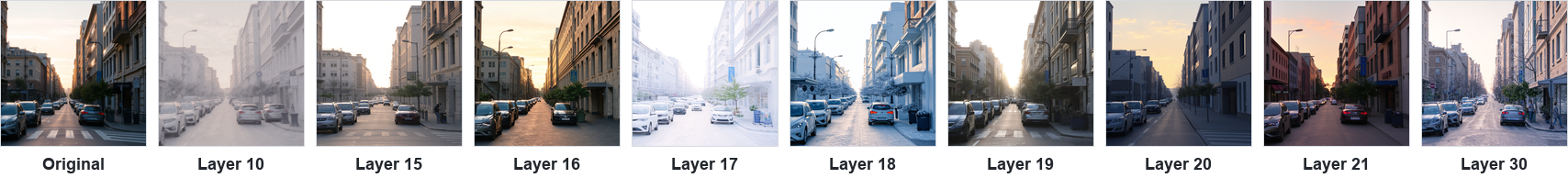}
    {A city street in the afternoon.}

    \vspace{0.8em}

    \QualAblationPanel
    {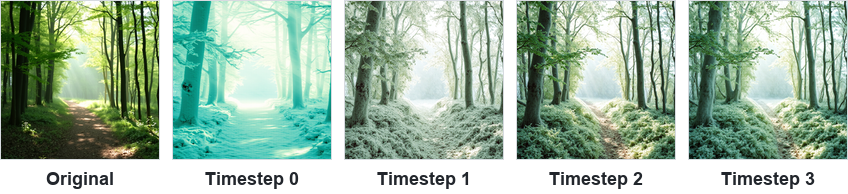}
    {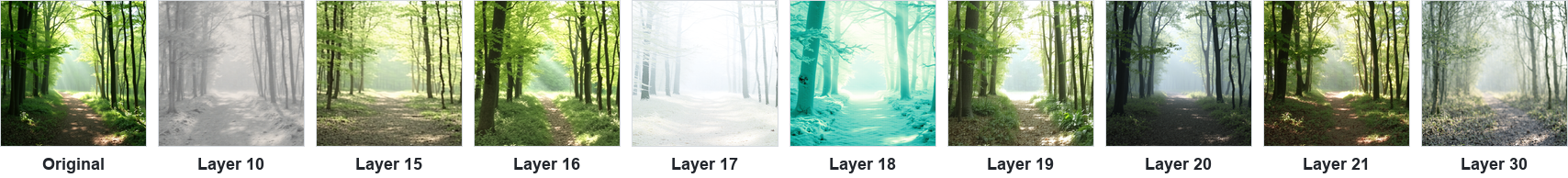}
    {A forest path in mild daylight.}

    \vspace{0.8em}

    \QualAblationPanel
    {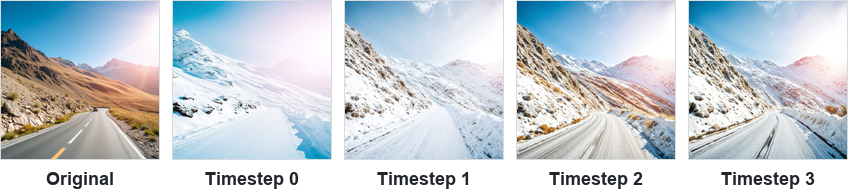}
    {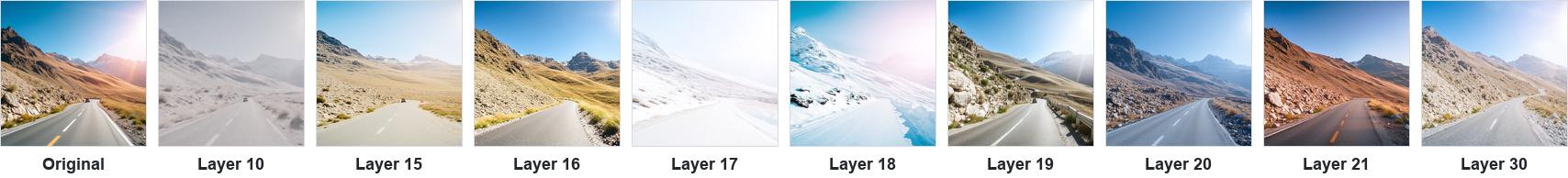}
    {A mountain road under bright daylight.}

    \caption{
    \textbf{Qualitative ablation samples -- global concept: \textit{snowy}.}
    Top row of each panel: extraction-timestep ablation. Bottom row: extraction-layer ablation.
    }
    \label{fig:qual_ablation_snowy}
\end{figure}

\Needspace{18\baselineskip}
\section{Prompt-Pair Sensitivity}
\label{app:prompt_sensitivity}

To evaluate the sensitivity of our method to prompt quality and concept-pair
design, we repeat the evaluation using substantially simpler positive/negative
prompt pairs. In contrast to our default prompts, these simplified pairs remove
carefully specified compositions, backgrounds, and target prominence, while
keeping the extraction and inference settings unchanged.

\begin{table}[H]
    \centering
    \caption{
    \textbf{Prompt-pair sensitivity ablation.}
    We evaluate our method using substantially simplified positive/negative
    prompt pairs while keeping all extraction and inference settings fixed.
    }
    \label{tab:prompt_sensitivity}
    \vspace{0.3em}
    \small
    \begin{tabular}{lcc}
        \toprule
        Category
        & $\Delta$VQA $\uparrow$
        & LPIPS $\downarrow$ \\
        \midrule
        Face   & 0.213 & 0.126 \\
        Object & 0.303 & 0.189 \\
        Global & 0.518 & 0.510 \\
        \midrule
        Average & \textbf{0.345} & \textbf{0.275} \\
        \bottomrule
    \end{tabular}
\end{table}

As shown in Table~\ref{tab:prompt_sensitivity}, substantially simplifying the
positive/negative prompt pairs does not degrade performance. Averaged across
the three concept families, simplified prompts achieve a $\Delta$VQA of 0.345
and an LPIPS of 0.275, compared with 0.32 and 0.29, respectively, for FLUX.1-schnell 5-pair setting. Despite removing carefully specified
compositions, backgrounds, and target prominence, performance remains
comparable and even slightly improves on average. This indicates that our
method is robust to prompt design and does not require carefully engineered
prompt pairs to extract effective concept directions.

\section{Failure Cases}
\label{app:failure_cases}

\begin{figure}[!htbp]
    \centering
    \includegraphics[width=0.5\linewidth]{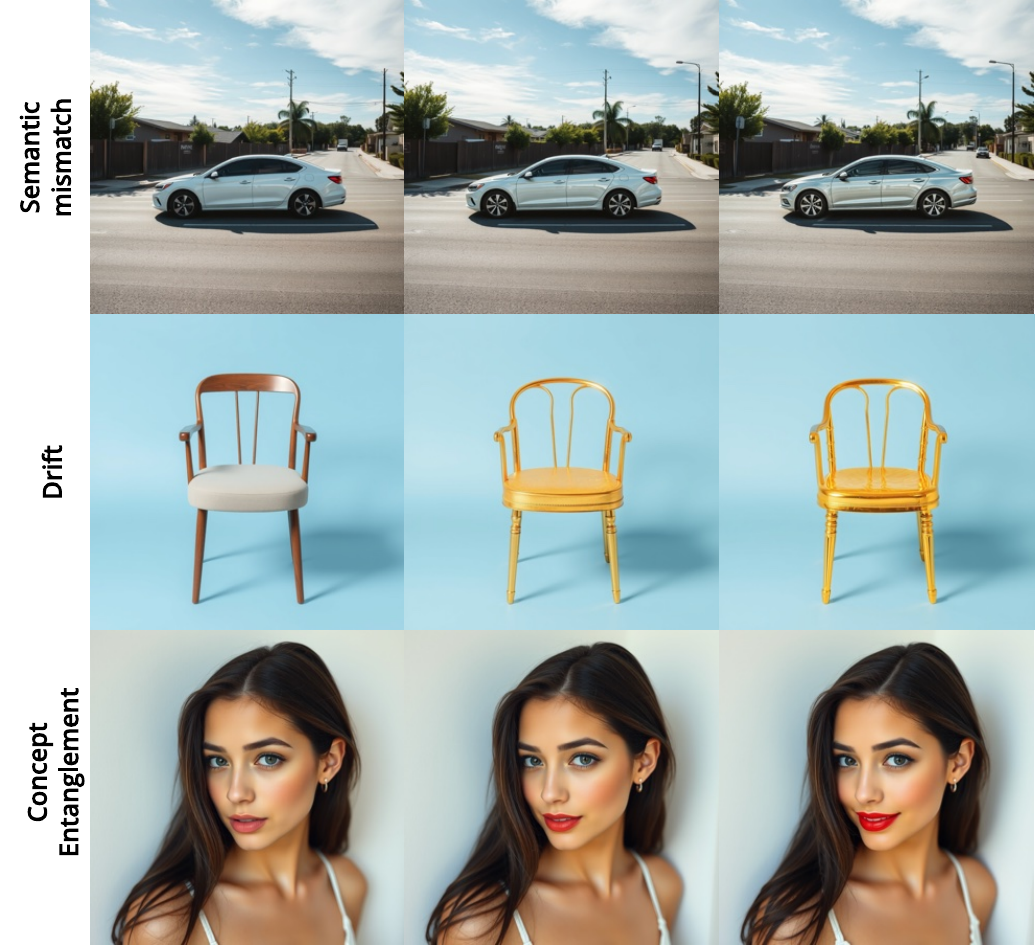}
    \caption{
    \textbf{Representative failure cases.}
    We show examples of semantic mismatch, spatial drift, and concept entanglement. In each row, increasing the steering strength introduces the target concept to different degrees, but may also produce weak edits, unwanted geometry changes, or correlated visual attributes.
    }
    \label{fig:failure_cases_appendix}
\end{figure}

Figure~\ref{fig:failure_cases_appendix} shows representative failure cases of our method. We observe three main types of failures. First, \textit{semantic mismatch} occurs when the extracted direction is applied in a setting where the target concept is not naturally supported by the image content (e.g. stripes but only on the reflection of the car). In this case, increasing the steering strength may produce only a weak or ambiguous edit rather than a coherent transformation. Second, \textit{drift} occurs when the target concept is introduced but the object geometry or spatial structure changes as well. This can happen for strong localized edits, where the steering direction affects not only the intended attribute but also the boundaries, pose, or shape of the object (e.g. changing the shape of the chair). Third, \textit{concept entanglement} occurs when the target direction also activates correlated attributes. For example, a facial direction may successfully introduce the desired attribute while also changing expression or other facial details (e.g. adding smile when injecting the lipstick concept). Disentanglement remains a long-standing challenge in generative modeling \cite{Wang_2020_CVPR}.

\Needspace{16\baselineskip}
\section{Runtime of Concept Extraction}
\label{app:runtime_concept_extraction}

We report the runtime required to extract concept directions from five positive-negative prompt pairs. These timings focus on the core concept-extraction pass. The benchmark uses three representative concepts: one object concept, one global concept, and one face concept. All times are reported in seconds. All experiments were run on a single NVIDIA H100 GPU.

\begin{table}[H]
\centering
\caption{
\textbf{Runtime for concept extraction and generation.}
For each concept, we report the total time to extract the concept direction from five positive-negative prompt pairs, the total time to generate five original images and five steered images, and the combined runtime. Generation time includes the steering-mask computation when applicable.
}
\label{tab:runtime_extraction_generation}
\scriptsize
\begin{tabular}{l l c c c c}
\toprule
Concept & Family & Extraction pairs & Extraction time & Generation time & Total time \\
\midrule
\textit{blue} & Object & 5 & 14.26s & 11.73s & 26.00s \\
\textit{cartoon} & Global & 5 & 10.52s & 10.73s & 21.25s \\
\textit{smile} & Face & 5 & 12.50s & 11.75s & 24.25s \\
\bottomrule
\end{tabular}
\end{table}

\Needspace{16\baselineskip}
\section{Outside-Mask LPIPS Evaluation}
\label{app:lpips_outside}

LPIPS measures visual drift from the original image, but does not indicate
whether changes occur within or outside the target region. We therefore
evaluate Outside-mask LPIPS, which measures changes outside the target region,
for face and object concepts at the maximum evaluated edit strength of each
method. For all methods, we use the same SAM~3 mask of the target region.
Lower values indicate less change outside the target region.

\begin{table}[H]
\centering
\small
\caption{\textbf{Outside mask LPIPS evaluation.} Lower values indicate less change outside the target region.}
\label{tab:outside_mask_lpips}
\begin{tabular}{lcc}
\toprule
Method & Face Outside mask LPIPS $\downarrow$ & Object Outside mask LPIPS $\downarrow$ \\
\midrule
Ours & 0.134 & 0.144 \\
FluxSpace & 0.055 & 0.124 \\
SAEdit & 0.153 & 0.250 \\
SliderEdit & 0.398 & 0.288 \\
Kontinuous Kontext & 0.290 & 0.202 \\
\bottomrule
\end{tabular}
\end{table}

Our method obtains the second-lowest Outside-mask LPIPS for both face and
object concepts, with 0.134 for faces and 0.144 for objects. It produces less
change outside the target region than SAEdit, and substantially less than
SliderEdit and Kontinuous Kontext. FluxSpace obtains lower values, but its edit
success is substantially weaker, with $\Delta$VQA of 0.17 compared with 0.32
for our method. These results show that, even under the strongest evaluated
steering, our method keeps changes largely within the target region while
achieving substantially stronger concept control than training-free baselines.

\Needspace{24\baselineskip}
\section{Robustness to Mask Perturbations}
\label{app:mask_perturbation}

Our method relies on spatial masks during concept extraction. To evaluate its
robustness to imperfect segmentation, we perturb the SAM~3 extraction masks using
erosion, dilation, spatial shifts, and increasing amounts of false-positive
regions. \textit{Erosion} shrinks the mask and removes part of the target region,
while \textit{dilation} expands it to include surrounding areas. \textit{Spatial
shifts} translate the mask away from its original location, simulating
localization errors. Finally, \textit{false positives} add unrelated regions to
the mask, introducing irrelevant spatial support during concept extraction.
We perform this analysis on face and object concepts and compare against the
original unperturbed masks.

\begin{table}[H]
    \centering
    \caption{
    \textbf{Robustness to extraction-mask perturbations.}
    We simulate segmentation errors through erosion, dilation, spatial shifts,
    and false-positive regions. Higher $\Delta$VQA indicates stronger editing,
    while lower LPIPS indicates better preservation.
    }
    \label{tab:mask_perturbation}
    \vspace{0.3em}
    \small
    \begin{tabular}{lcccc}
        \toprule
        Extraction-mask perturbation
        & Face $\Delta$VQA $\uparrow$
        & Face LPIPS $\downarrow$
        & Object $\Delta$VQA $\uparrow$
        & Object LPIPS $\downarrow$ \\
        \midrule
        Erosion -- small          & 0.197 & 0.267 & 0.294 & 0.179 \\
        Erosion -- medium         & 0.134 & 0.228 & 0.280 & 0.169 \\
        Dilation -- small         & 0.188 & 0.266 & 0.301 & 0.185 \\
        Dilation -- medium        & 0.191 & 0.256 & 0.305 & 0.190 \\
        Shift -- small            & 0.194 & 0.241 & 0.290 & 0.180 \\
        Shift -- medium           & 0.189 & 0.233 & 0.288 & 0.181 \\
        False positives -- 25\%   & 0.186 & 0.247 & 0.309 & 0.180 \\
        False positives -- 50\%   & 0.182 & 0.251 & 0.299 & 0.187 \\
        False positives -- 75\%   & 0.149 & 0.235 & 0.300 & 0.186 \\
        \midrule
        Unperturbed               & 0.185 & 0.189 & 0.287 & 0.164 \\
        \bottomrule
    \end{tabular}
\end{table}

As shown in Table~\ref{tab:mask_perturbation}, concept extraction remains
effective under substantial mask perturbations. Object concepts are particularly
robust, with $\Delta$VQA remaining between 0.280 and 0.309 across every
perturbation. Face concepts are more sensitive because their relevant attributes
occupy smaller spatial regions: medium erosion and 75\% false positives reduce
face $\Delta$VQA to 0.134 and 0.149, respectively, whereas small perturbations
and spatial shifts largely preserve edit effectiveness. Perturbations generally
increase LPIPS relative to the unperturbed masks, showing that inaccurate masks
primarily introduce additional visual drift rather than preventing the requested
edit. Dilation and moderate false positives can even strengthen $\Delta$VQA, but
at the cost of higher LPIPS, suggesting that including additional tokens
produces a stronger yet less spatially specific direction. Overall, these
results demonstrate that our method does not require perfect segmentation and
tolerates moderate mask errors, particularly for object concepts, while accurate
spatial support remains beneficial for preserving unrelated content and
extracting fine-grained facial attributes.

\section{Additional Preservation Metrics}
\label{app:additional_metrics}

We further evaluate content and identity preservation using two additional
metrics: DINO similarity and face cosine similarity. For our method, we report
results using FLUX.1-schnell with 5 positive/negative prompt pairs, consistent
with our default evaluation setting. Higher values indicate better preservation.

\begin{table}[H]
    \centering

    \begin{minipage}[t]{0.48\textwidth}
        \centering
        \small
        \captionof{table}{
        \textbf{DINO similarity.}
        Higher similarity indicates stronger preservation of the original
        image representation.
        }
        \label{tab:dino_similarity}
        \vspace{0.3em}

        \begin{tabular}{lc}
            \toprule
            Method & DINO Sim. $\uparrow$ \\
            \midrule
            K. Kontext & 0.874 \\
            SliderEdit & 0.815 \\
            SAEdit     & 0.764 \\
            FluxSpace  & \textbf{0.877} \\
            Ours (Schnell, 5 pairs) & 0.741 \\
            \bottomrule
        \end{tabular}
    \end{minipage}
    \hfill
    \begin{minipage}[t]{0.48\textwidth}
        \centering
        \small
        \captionof{table}{
        \textbf{Face similarity.}
        Mean cosine similarity measures identity preservation for face edits.
        Higher is better.
        }
        \label{tab:face_similarity}
        \vspace{0.3em}

        \begin{tabular}{lc}
            \toprule
            Method & Face Sim. $\uparrow$ \\
            \midrule
            K. Kontext & 0.791 \\
            SliderEdit & 0.762 \\
            SAEdit     & 0.797 \\
            FluxSpace  & \textbf{0.926} \\
            Ours (Schnell, 5 pairs) & 0.748 \\
            \bottomrule
        \end{tabular}
    \end{minipage}

\end{table}

As shown in Tables~\ref{tab:dino_similarity} and~\ref{tab:face_similarity},
our method obtains lower preservation scores than the baselines. These metrics
compare global image and face representations, so they decrease with any
visible edit, including the requested one. Attributes such as \textit{old},
\textit{beard}, or \textit{glasses} directly alter face embeddings, and global
concepts such as \textit{anime} or \textit{snowy} alter the whole-image DINO
representation. Methods with weaker edits therefore naturally obtain higher
similarity. FluxSpace obtains the highest DINO and face similarities while
reaching only 0.17 $\Delta$VQA, compared with 0.32 for our method. K. Kontext
and SliderEdit additionally use the reference image during editing and are
trained for this task. Measured outside the target region
(Appendix~\appref{app:lpips_outside}), our method preserves unrelated content
better than all baselines except FluxSpace.

\section{Real-Image Steering with FLUX.2}
\label{app:flux2_real_image}

We provide a qualitative proof of concept for real-image steering using
FLUX.2-Klein 4B. Concept directions are extracted exclusively from synthetic
positive/negative prompt pairs and then applied to real reference images.
The conditioning prompt is left empty, all inputs are kept fixed, and only
the steering strength $\alpha$ is varied.

\begin{figure}[!htbp]
    \centering
    \includegraphics[width=0.95\textwidth]{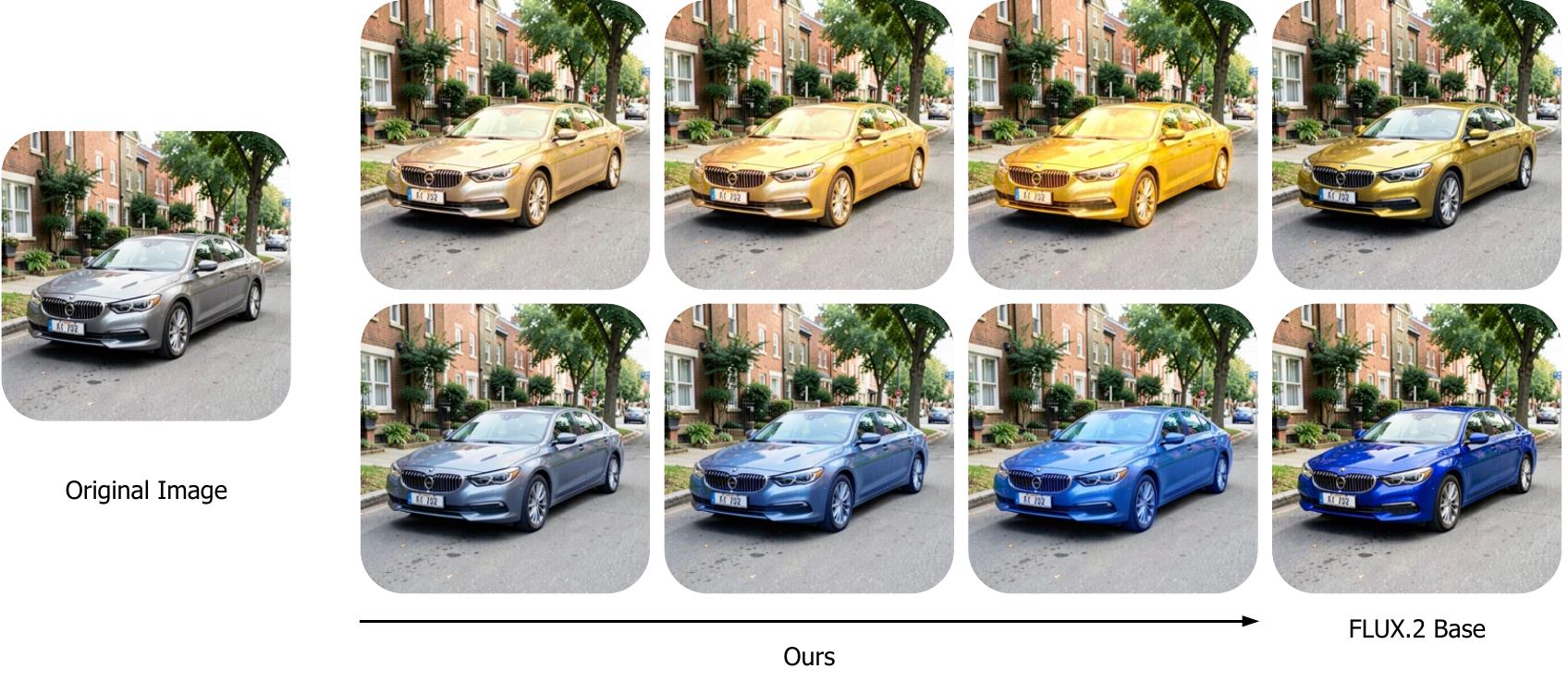}
    \caption{
    \textbf{Real-image steering with FLUX.2.}
    Concept directions extracted exclusively from synthetic prompt pairs
    transfer to real reference images across multiple concepts. For each
    example, increasing the steering strength $\alpha$ progressively increases
    the presence of the target concept while preserving the reference image.
    In contrast to FLUX.2 direct prompt editing, which produces a single edit
    for a given instruction, our approach provides continuous and progressive
    control over edit strength.
    }
    \label{fig:flux2_real_image}
\end{figure}

Following the same relative configuration used for FLUX.1, we extract the
direction from block 7 at timestep 0 and inject it over blocks 7--24. As shown
in Figure~\ref{fig:flux2_real_image}, directions learned from synthetic prompt
pairs transfer successfully to real-image editing across multiple concepts, namely "gold" and "blue" here.
Increasing $\alpha$ progressively strengthens the presence of the target concept
while preserving the structure and content of the reference image.

An important advantage over FLUX.2's native prompt-based editing is the
continuous control provided by activation steering. While direct prompt editing
produces a single edited output for a given instruction, our method allows the
target concept to be introduced with a progressive increase in strength as
$\alpha$ increases.

\FloatBarrier
\section{More Qualitative Samples}
\label{app:more_qual}

\begin{figure}[H]
    \centering
    \includegraphics[height=0.78\textheight,keepaspectratio]{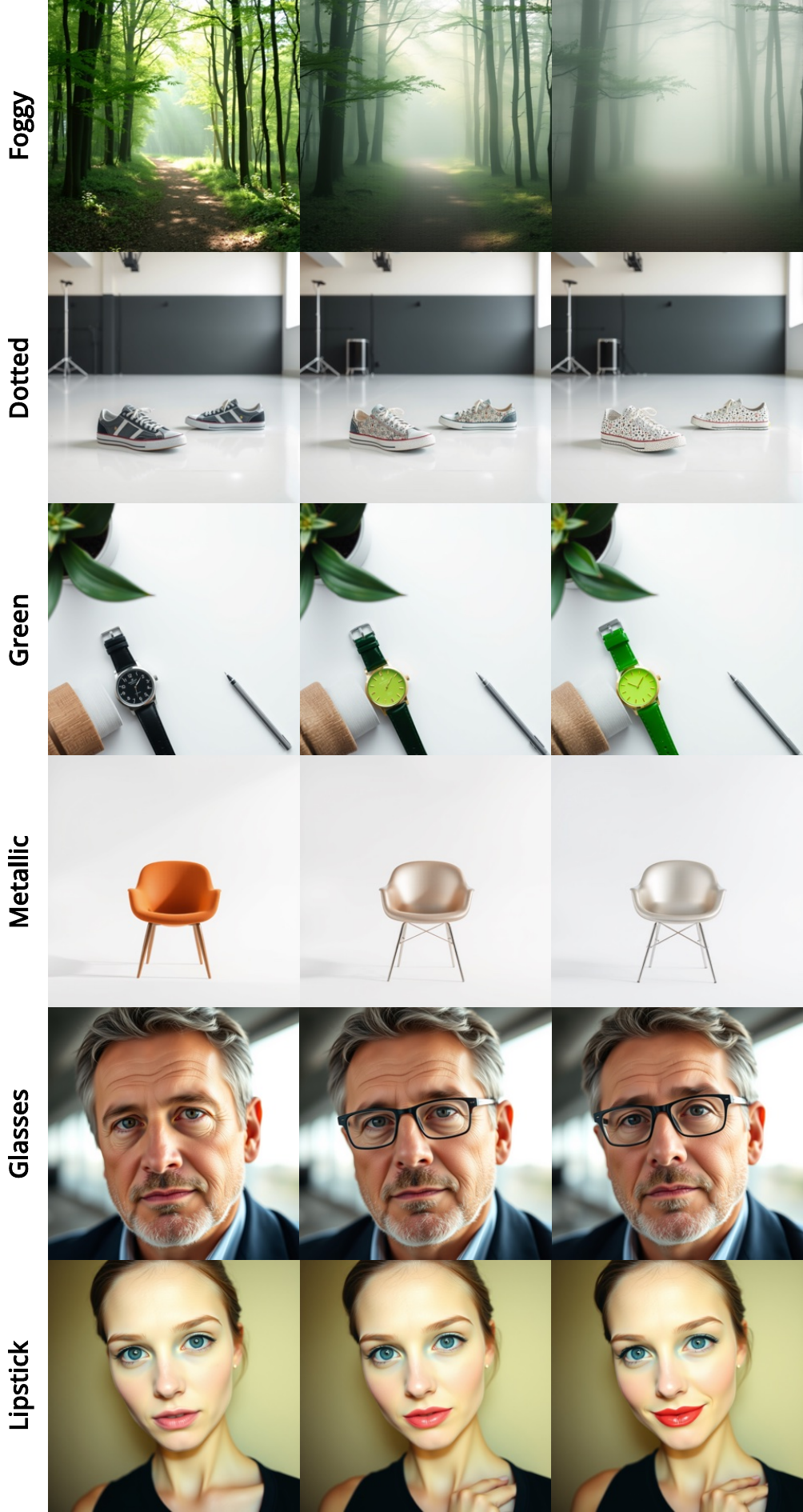}
    \caption{
    \textbf{Qualitative examples of concept steering.}
    We present more qualitative samples across concepts.
    }
    \label{fig:more_qual}
\end{figure}


\newpage
\section*{NeurIPS Paper Checklist}

\begin{enumerate}

\item {\bf Claims}
    \item[] Question: Do the main claims made in the abstract and introduction accurately reflect the paper's contributions and scope?
    \item[] Answer: \answerYes{}
    \item[] Justification: The abstract and Introduction state the main claims, namely, few-shot, training-free concept extraction from DiT activations, masked difference-of-means for localized concepts and inference-time global, local, and compositional steering. The scope of the claims is supported by the Sections Method, Experiments, Ablation, and Failure Cases.

\item {\bf Limitations}
    \item[] Question: Does the paper discuss the limitations of the work performed by the authors?
    \item[] Answer: \answerYes{}
    \item[] Justification: The paper discusses limitations in Appendix~\appref{app:failure_cases} and Section~\ref{sec:conclusion}, including drift at high steering strengths, semantic mismatch, and concept entanglement.
    \item[] Guidelines:
    \begin{itemize}
        \item The answer \answerNA{} means that the paper has no limitation while the answer \answerNo{} means that the paper has limitations, but those are not discussed in the paper. 
        \item The authors are encouraged to create a separate ``Limitations'' section in their paper.
        \item The paper should point out any strong assumptions and how robust the results are to violations of these assumptions.
        \item The authors should reflect on the scope of the claims made, e.g., if the approach was only tested on a few datasets or with a few runs.
        \item The authors should reflect on the factors that influence the performance of the approach.
        \item The authors should discuss the computational efficiency of the proposed algorithms and how they scale with dataset size.
        \item If applicable, the authors should discuss possible limitations of their approach to address problems of privacy and fairness.
        \item Reviewers will be specifically instructed to not penalize honesty concerning limitations.
    \end{itemize}

\item {\bf Theory assumptions and proofs}
    \item[] Question: For each theoretical result, does the paper provide the full set of assumptions and a complete (and correct) proof?
    \item[] Answer: \answerNA{}
    \item[] Justification: The paper does not present theoretical results or formal theorems. The mathematical expressions in the Method Section define the estimator and inference-time intervention rather than state theoretical guarantees.

\item {\bf Experimental result reproducibility}
    \item[] Question: Does the paper fully disclose all the information needed to reproduce the main experimental results of the paper to the extent that it affects the main claims and/or conclusions of the paper (regardless of whether the code and data are provided or not)?
    \item[] Answer: \answerYes{}
    \item[] Justification: The paper describes the model, extraction layer, extraction timestep, injection range, evaluation concepts, steering strengths, metrics, and baselines in Sections~\ref{sec:method} and~\ref{sec:experiments}. Additional prompt, metric, baseline, and runtime details are provided in the appendix.
    \item[] Guidelines:
    \begin{itemize}
        \item The answer \answerNA{} means that the paper does not include experiments.
        \item If the paper includes experiments, a \answerNo{} answer to this question will not be perceived well by the reviewers.
        \item If the contribution is a dataset and\slash or model, the authors should describe the steps taken to make their results reproducible or verifiable. 
        \item Depending on the contribution, reproducibility can be accomplished in various ways.
        \item While NeurIPS does not require releasing code, the conference does require all submissions to provide some reasonable avenue for reproducibility, which may depend on the nature of the contribution.
        \begin{enumerate}
            \item If the contribution is primarily a new algorithm, the paper should make it clear how to reproduce that algorithm.
            \item If the contribution is primarily a new model architecture, the paper should describe the architecture clearly and fully.
            \item If the contribution is a new model, then there should either be a way to access this model for reproducing the results or a way to reproduce the model.
            \item In the case of closed-source models, it may be that access to the model is limited in some way, but it should be possible for other researchers to have some path to reproducing or verifying the results.
        \end{enumerate}
    \end{itemize}

\item {\bf Open access to data and code}
    \item[] Question: Does the paper provide open access to the data and code, with sufficient instructions to faithfully reproduce the main experimental results, as described in supplemental material?
    \item[] Answer: \answerYes{}
    \item[] Justification: The paper provides implementation and evaluation details sufficient to reproduce the main experiments. The link to the code is available on the project page: \href{https://newbyl.github.io/few_shot_project_page/}{\nolinkurl{newbyl.github.io/few_shot_project_page}}.
    \item[] Guidelines:
    \begin{itemize}
        \item The answer \answerNA{} means that paper does not include experiments requiring code.
        \item Please see the NeurIPS code and data submission guidelines (\url{https://neurips.cc/public/guides/CodeSubmissionPolicy}) for more details.
        \item While we encourage the release of code and data, we understand that this might not be possible, so \answerNo{} is an acceptable answer.
        \item The instructions should contain the exact command and environment needed to run to reproduce the results.
        \item The authors should provide instructions on data access and preparation.
        \item The authors should provide scripts to reproduce all experimental results for the new proposed method and baselines.
        \item At submission time, to preserve anonymity, the authors should release anonymized versions, if applicable.
        \item Providing as much information as possible in supplemental material is recommended, but including URLs to data and code is permitted.
    \end{itemize}

\item {\bf Experimental setting/details}
    \item[] Question: Does the paper specify all the training and test details (e.g., data splits, hyperparameters, how they were chosen, type of optimizer) necessary to understand the results?
    \item[] Answer: \answerYes{}
    \item[] Justification: The paper specifies the main experimental settings in Section~\ref{sec:experiments}, including the base model, extraction layer, extraction timestep, injection range, evaluated concepts, steering strengths, and metrics. Additional details about prompts, baselines, and metric implementation are provided in the appendix.

\item {\bf Experiment statistical significance}
    \item[] Question: Does the paper report error bars suitably and correctly defined or other appropriate information about the statistical significance of the experiments?
    \item[] Answer: \answerNo{}
    \item[] Justification: We report aggregate metrics over prompts, concepts, seeds, and steering strengths, but the main tables do not include error bars or formal statistical significance tests. The evaluation uses multiple prompts and seeds to reduce dependence on individual samples.
    \item[] Guidelines:
    \begin{itemize}
        \item The answer \answerNA{} means that the paper does not include experiments.
        \item The authors should answer \answerYes{} if the results are accompanied by error bars, confidence intervals, or statistical significance tests, at least for the experiments that support the main claims of the paper.
        \item The factors of variability that the error bars are capturing should be clearly stated.
        \item The method for calculating the error bars should be explained.
        \item The assumptions made should be given.
        \item It should be clear whether the error bar is the standard deviation or the standard error of the mean.
        \item It is OK to report 1-sigma error bars, but one should state it.
        \item For asymmetric distributions, the authors should be careful not to show symmetric error bars that would yield results that are out of range.
        \item If error bars are reported in tables or plots, the authors should explain in the text how they were calculated and reference the corresponding figures or tables.
    \end{itemize}

\item {\bf Experiments compute resources}
    \item[] Question: For each experiment, does the paper provide sufficient information on the computer resources (type of compute workers, memory, time of execution) needed to reproduce the experiments?
    \item[] Answer: \answerYes{}
    \item[] Justification: The paper and appendix report the relevant model and runtime settings used for evaluation. In particular, the runtime appendix reports concept-extraction timings and the GPU used for this work.
    \item[] Guidelines:
    \begin{itemize}
        \item The answer \answerNA{} means that the paper does not include experiments.
        \item The paper should indicate the type of compute workers CPU or GPU, internal cluster, or cloud provider, including relevant memory and storage.
        \item The paper should provide the amount of compute required for each of the individual experimental runs as well as estimate the total compute. 
        \item The paper should disclose whether the full research project required more compute than the experiments reported in the paper.
    \end{itemize}

\item {\bf Code of ethics}
    \item[] Question: Does the research conducted in the paper conform, in every respect, with the NeurIPS Code of Ethics \url{https://neurips.cc/public/EthicsGuidelines}?
    \item[] Answer: \answerYes{}
    \item[] Justification: The research conforms to the NeurIPS Code of Ethics. The work studies controllable image generation using generated images and released models, and does not involve private data, human-subject experiments, or deployment on users.
    \item[] Guidelines:
    \begin{itemize}
        \item The answer \answerNA{} means that the authors have not reviewed the NeurIPS Code of Ethics.
        \item If the authors answer \answerNo, they should explain the special circumstances that require a deviation from the Code of Ethics.
        \item The authors should make sure to preserve anonymity.
    \end{itemize}

\item {\bf Broader impacts}
    \item[] Question: Does the paper discuss both potential positive societal impacts and negative societal impacts of the work performed?
    \item[] Answer: \answerYes{}
    \item[] Justification: The paper concerns image steering, which can improve controllability and interpretability of generative models, but may also make image manipulation easier. We discuss limitations and failure modes, and the method is evaluated as a research prototype rather than a deployed system.
    \item[] Guidelines:
    \begin{itemize}
        \item The answer \answerNA{} means that there is no societal impact of the work performed.
        \item If the authors answer \answerNA{} or \answerNo, they should explain why their work has no societal impact or why the paper does not address societal impact.
        \item Examples of negative societal impacts include potential malicious or unintended uses, fairness considerations, privacy considerations, and security considerations.
        \item The conference expects that many papers will be foundational research and not tied to particular applications, let alone deployments.
        \item The authors should consider possible harms that could arise when the technology is used as intended, when it gives incorrect results, and from misuse.
        \item If there are negative societal impacts, the authors could also discuss possible mitigation strategies.
    \end{itemize}

\item {\bf Safeguards}
    \item[] Question: Does the paper describe safeguards that have been put in place for responsible release of data or models that have a high risk for misuse (e.g., pre-trained language models, image generators, or scraped datasets)?
    \item[] Answer: \answerNA{}
    \item[] Justification: We do not release a new high-risk pretrained generative model or scraped dataset. The method operates on existing released image-generation models and extracts lightweight concept directions for research evaluation.
    \item[] Guidelines:
    \begin{itemize}
        \item The answer \answerNA{} means that the paper poses no such risks.
        \item Released models that have a high risk for misuse or dual-use should be released with necessary safeguards to allow for controlled use of the model.
        \item Datasets that have been scraped from the Internet could pose safety risks. The authors should describe how they avoided releasing unsafe images.
        \item We recognize that providing effective safeguards is challenging, and many papers do not require this, but we encourage authors to take this into account and make a best faith effort.
    \end{itemize}

\item {\bf Licenses for existing assets}
    \item[] Question: Are the creators or original owners of assets (e.g., code, data, models), used in the paper, properly credited and are the license and terms of use explicitly mentioned and properly respected?
    \item[] Answer: \answerYes{}
    \item[] Justification: The paper cites the existing models, methods, and metrics used in the experiments, including FLUX, SAM, VQAScore, LPIPS, and the compared baselines. We use released assets according to their intended research usage.
    \item[] Guidelines:
    \begin{itemize}
        \item The answer \answerNA{} means that the paper does not use existing assets.
        \item The authors should cite the original paper that produced the code package or dataset.
        \item The authors should state which version of the asset is used and, if possible, include a URL.
        \item The name of the license should be included for each asset.
        \item For scraped data from a particular source, the copyright and terms of service of that source should be provided.
        \item If assets are released, the license, copyright information, and terms of use in the package should be provided.
        \item For existing datasets that are re-packaged, both the original license and the license of the derived asset should be provided.
        \item If this information is not available online, the authors are encouraged to reach out to the asset's creators.
    \end{itemize}

\item {\bf New assets}
    \item[] Question: Are new assets introduced in the paper well documented and is the documentation provided alongside the assets?
    \item[] Answer: \answerNA{}
    \item[] Justification: The paper does not introduce or release a new dataset, pretrained model, or benchmark asset. The concept prompts and evaluation details are documented in the appendix for reproducibility.
    \item[] Guidelines:
    \begin{itemize}
        \item The answer \answerNA{} means that the paper does not release new assets.
        \item Researchers should communicate the details of the dataset\slash code\slash model as part of their submissions via structured templates.
        \item The paper should discuss whether and how consent was obtained from people whose asset is used.
        \item At submission time, remember to anonymize your assets, if applicable.
    \end{itemize}

\item {\bf Crowdsourcing and research with human subjects}
    \item[] Question: For crowdsourcing experiments and research with human subjects, does the paper include the full text of instructions given to participants and screenshots, if applicable, as well as details about compensation (if any)? 
    \item[] Answer: \answerNA{}
    \item[] Justification: The paper does not involve crowdsourcing experiments or research with human subjects. All evaluation images are generated synthetically from text prompts.

\item {\bf Institutional review board (IRB) approvals or equivalent for research with human subjects}
    \item[] Question: Does the paper describe potential risks incurred by study participants, whether such risks were disclosed to the subjects, and whether Institutional Review Board (IRB) approvals (or an equivalent approval/review based on the requirements of your country or institution) were obtained?
    \item[] Answer: \answerNA{}
    \item[] Justification: The paper does not involve human-subject experiments, crowdsourcing, or participant data, so IRB approval or equivalent review is not applicable.
    \item[] Guidelines:
    \begin{itemize}
        \item The answer \answerNA{} means that the paper does not involve crowdsourcing nor research with human subjects.
        \item Depending on the country in which research is conducted, IRB approval or equivalent may be required for any human subjects research.
        \item We recognize that the procedures for this may vary significantly between institutions and locations, and we expect authors to adhere to the NeurIPS Code of Ethics and the guidelines for their institution. 
        \item For initial submissions, do not include any information that would break anonymity, if applicable.
    \end{itemize}

\item {\bf Declaration of LLM usage}
    \item[] Question: Does the paper describe the usage of LLMs if it is an important, original, or non-standard component of the core methods in this research? Note that if the LLM is used only for writing, editing, or formatting purposes and does \emph{not} impact the core methodology, scientific rigor, or originality of the research, declaration is not required.
    \item[] Answer: \answerYes{}
    \item[] Justification: We declare LLM usage for the positive and negative pairs generation. Beyond this, any language-model assistance, if used, was limited to writing, editing, or formatting and did not affect the scientific methodology.

\end{enumerate}

\end{document}